\documentclass[twoside,11pt]{article}

\usepackage{jmlr2e}
 \usepackage{amsmath} 
\usepackage{amsfonts} 
\usepackage{graphics}
\usepackage{color}
\usepackage[space]{grffile}
\usepackage{algorithmic}
\usepackage{algorithm}
\usepackage{tikz}
\usepackage{diagbox}

\newcommand{\fracpartial}[2]{\frac{\partial {#1}}{\partial {#2}}}

\newcommand{\vecT}[0]{\vec{\tau}}
\newcommand{\vecW}[0]{\vec{w}}
\newcommand{\vecA}[0]{\vec{a}}

\newcommand{\vecX}[0]{\vec{x}}
\newcommand{\vecY}[0]{\vec{y}}
\newcommand{\vecAllX}[0]{\{\vec{x}_1, \vec{x}_2, \ldots, \vec{x}_\np\}} 
\newcommand{\vecAllA}[0]{\{\vec{a}_1, \vec{a}_2, \ldots, \vec{a}_\np\}}

\newcommand{\errorFunction}{L}
\newcommand{\errorFunctionT}{L'}

\newcommand{\dEdt}[0]{\fracpartial{\errorFunctionT}{\vecT}}
\newcommand{\dEdw}[0]{\fracpartial{\errorFunction}{\vecW}}
\newcommand{\nl}[0]{{n_\mathrm{L}}}
\newcommand{\np}[0]{{n_\mathrm{p}}}
\newcommand{\nb}[0]{{n_\mathrm{b}}}
\newcommand{\nbt}[0]{{\overline{n}_\mathrm{b}}}
\newcommand{\nw}[0]{{n_\mathrm{w}}}
\newcommand{\cl}[0]{{c_\mathrm{L}}}
\newcommand{\tloop}[0]{{t}}
\newcommand{\nloop}{{n_\mathrm{t}}}
\newcommand{\nloopt}{{\overline{n}_\mathrm{t}}}

\newcommand{\norm}[1]{\left|\left|{#1}\right|\right|}
\newcommand{\hadamard}[2]{#1 \odot #2}
\newcommand{\kroneckerMatrix}[2]{[J^{#1 #2}]}
\newcommand{\Kij}[0]{\kroneckerMatrix{i}{j}}

\newcommand{\Kmn}[0]{\kroneckerMatrix{m}{n}}

\newcommand{\frobProduct}[2]{\left<#1,#2\right>_{\!F}}
\newcommand{\createMatrix}[2]{\left({#2}\right) ^{[#1]}}
\newcommand{\deltaStar}[1]{\delta{#1}}
\newcommand{\Bmn}[0]{B^{mn}}
\newcommand{\Reals}[0]{\mathbb{R}}
\newcommand{\isDefinedToBe}[0]{:=}
\newcommand{\optimal}[1]{{#1}^{*}}
\newcommand{\pseudoinv}[0]{\dagger} 
\newcommand{\NRpseudoinv}[0]{+} 
\newcommand{\pseudoinvinv}[0]{\ddagger} 

\newcommand{\ifIncludeMikesThoughts}[1]{}
\newcommand{\ifIncludeToDos}[1]{}
\newcommand{\todo}[1]{}
\newcommand{\hide}[1]{}

\newcommand{\ifIncludeProbenExpts}[1]{}

\newcommand{\ovA}{A}
\newcommand{\ovS}{S}
\newcommand{\layerInputStack}[1]{{A}_{{[0:{#1})}}}
\newcommand{\layerInputStackOv}[1]{{\ovA}_{{[0:{#1})}}}
\newcommand{\layerWeightStack}[1]{{W}_{{[0:{#1}]}}}
\DeclareMathOperator{\relu}{ReLU}
\DeclareMathOperator{\lrelu}{LReL}
\newcommand{\numinputs}{n_\mathrm{i}}
\newcommand{\numoutputs}{n_\mathrm{o}}
\newcommand{\numinputchannels}{n_\mathrm{ic}}
\newcommand{\numoutputchannels}{n_\mathrm{oc}}
\newcommand{\kernelwidth}{k_\mathrm{w}}
\newcommand{\kernelheight}{k_\mathrm{h}}
\newcommand{\sout}[1]{}
\newcommand{\mappingFunction}[0]{m}
\newcommand{\mappingFunctionFull}[0]{\mappingFunction(\overline{X},\vecT)}

\usepackage{lastpage}

\jmlrheading{22}{2021}{1-\pageref{LastPage}}{1/20; Revised
11/21}{12/21}{20-040}{Michael Fairbank, Spyros Samothrakis and Luca Citi}

\ShortHeadings{Deep Learning in Target Space}{Fairbank,  Samothrakis and Citi}
\firstpageno{1}

\begin{document}

\graphicspath{{resultgraphs/mnist/}{\detokenize{resultgraphs/twoSpirals/}}{resultgraphs/bitSequence/}{resultgraphs/imdb/}}
\title{Deep Learning in Target Space}

\author{\name Michael Fairbank \email m.fairbank@essex.ac.uk \\
 \name Spyridon Samothrakis \email ssamot@essex.ac.uk \\
 \name Luca Citi \email lciti@essex.ac.uk \\
\addr Department of Computer Science and Electronic Engineering\\
University of Essex\\
Colchester, CO4 3SQ, UK}

\editor{Miguel Carreira-Perpinan}

\maketitle

\begin{abstract}
Deep learning uses neural networks which are parameterised by their weights.  The neural networks are usually trained by tuning the weights to directly minimise a given loss function.  In this paper we propose to re-parameterise the weights into targets for the firing strengths of the individual nodes in the network. Given a set of targets, it is possible to calculate the weights which make the firing strengths best meet those targets. It is argued that using targets for training addresses the problem of exploding gradients, by a process which we call cascade untangling, and  makes the loss-function surface smoother to traverse, and so leads to easier, faster training, and also potentially better generalisation, of the neural network.  It also allows for easier learning of deeper and recurrent network structures. The necessary conversion of targets to weights comes at an extra computational expense, which is in many cases manageable.  Learning in target space can be combined with existing neural-network optimisers, for extra gain.  Experimental results show the speed of using target space, and examples of improved generalisation, for fully-connected networks and convolutional networks, and the ability to recall and process long time sequences and perform natural-language processing with recurrent networks.  
\end{abstract}


\section{Introduction} \label{sec:introduction}

\begin{figure}[ht]
\begin{center} 
\def\layersep{1.5cm}
\usetikzlibrary{decorations.pathreplacing}
\begin{tikzpicture}
    [draw=black!80, node distance=\layersep]
    \tikzstyle{neuron}=[circle,draw=black,thick, minimum size=12pt]
    \tikzstyle{input neuron}=[neuron];
    \tikzstyle{output neuron}=[neuron];
    \tikzstyle{hidden neuron}=[neuron, fill=black!20];
    
    \foreach \name / \y in {1,...,3}
        \node[input neuron] (I-\name) at (0,-\y) {};

    \foreach \name / \y in {1,...,2}
        \path[yshift=-0.5cm]
            node[hidden neuron] (H-\name) at (\layersep,-\y cm) {};
    
    \foreach \name / \y in {1,...,3}
        \path[yshift=0cm]
            node[hidden neuron, right of=H-1] (J-\name) at (\layersep,-\y cm) {};
    \foreach \name / \y in {1,...,2}
        \path[yshift=-0.5cm]
            node[hidden neuron, right of=J-2] (K-\name) at (\layersep*2,-\y cm) {};
    
    \foreach \name / \y in {1,...,3}
       \path[yshift=0cm] node[output neuron, right of=K-1] (O-\name) at (\layersep*3,-\y cm) {};

    \foreach \source in {1,...,3}
        \foreach \dest in {1,...,2}
            \path  (I-\source) edge[->,shorten >=1pt] (H-\dest);

    \foreach \source in {1,...,2}
        \foreach \dest in {1,...,3}
            \path (H-\source) edge[->,shorten >=1pt] (J-\dest);
            
    \foreach \source in {1,...,3}
        \foreach \dest in {1,...,2}
            \path (J-\source) edge[->,shorten >=1pt] (K-\dest);
            
    \foreach \source in {1,...,2}
        \foreach \dest in {1,...,3}
            \path (K-\source) edge[->,shorten >=1pt] (O-\dest);

    \draw [decorate,decoration={brace,amplitude=10pt},xshift=-4pt,yshift=0pt] (-0.2,-3.2) -- (-0.2,-0.8) node [black,midway,xshift=-0.6cm,rotate=90] {\small Input $\vecX$};
    \draw [decorate,decoration={brace,amplitude=10pt,mirror,raise=4pt},yshift=0pt] (6.2,-3.2) -- (6.2,-0.8) node [black,midway,xshift=0.8cm,rotate=90] {\small Output $\vecY$};
    \draw [decorate,decoration={brace,amplitude=5pt,mirror,raise=4ex}]
  (1.1,-3.1) -- (4.9,-3.1) node[midway,yshift=-3em]{\small Hidden Layers};
    \node[black,below of=I-3,shift={(-1, 1)}] (layer) {\small Layer:};
    \node[black,below of=I-3,shift={(0, 1)}] (layer) {\small 1};
    \node[black,below of=I-3,shift={(\layersep, 1)}] (layer) {\small 2};
    \node[black,below of=I-3,shift={(\layersep*2, 1)}] (layer) {\small 3};
    \node[black,below of=I-3,shift={(\layersep*3, 1)}] (layer) {\small 4};
    \node[black,below of=I-3,shift={(\layersep*4, 1)}] (layer) {\small 5};
\end{tikzpicture}
\end{center}      
\caption {Example feed-forward NN with structure ``3-2-3-2-3'', with five layers  ($\nl=5$).  An input vector $\vecX \in \Reals ^3$ (in this example) is fed in from the left.  Data propagates along the forward arrows (weights) causing nodes to fire, layer by layer, eventually producing output vector $\vecY \in \Reals ^3$.  The precise equations governing a NN are given in Section \ref{sec:NNDefinition}. Bias weights are not shown here, and this NN does not include shortcut connections.  
}
    \label{fig:ffnnSchematics}
\end{figure}
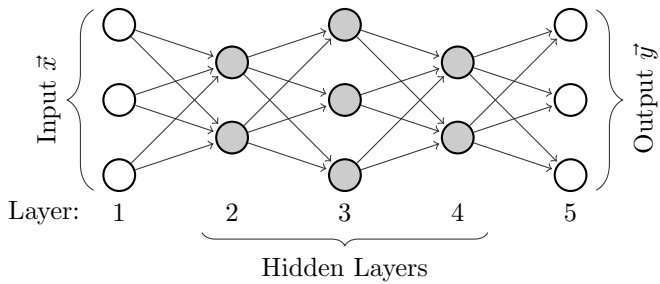

A feed-forward artificial neural network (NN) is a function $f(\vecX, \vecW)$, parameterised by a weights vector $\vecW$, that maps an input vector $\vecX$ to an output vector $\vecY=f(\vecX, \vecW)$.  This paper initially considers  feed-forward fully-connected layered NNs with $\nl$ layers, as illustrated in Figure \ref{fig:ffnnSchematics}.

NNs can be used in many problem domains, including pattern recognition, classification and function approximation \citep{bishop95,Goodfellow-et-al-2016}.  There are also numerous industrial and scientific applications for NNs, including vision, neurocontrol, language translation, image captioning, reinforcement learning and game playing  \citep{silver2017mastering,mnih2015human,karpathy2015deep,fairbank14clipping,Fairbank201474,sutskever2014sequence,samothrakis2016ijcnn}.

Training a NN means deciding upon an appropriate value for the weights vector $\vecW$ so that the NN performs the desired task successfully.
This training process is usually an iterative numerical method that works by trying continually to adjust $\vecW$ so as to minimise some real-valued loss function $\errorFunction(\vecX_1,\vecX_2, \ldots, \vecX_\np,\vecW)$ for a given set of $\np$ example input vectors $(\vecX_1,\vecX_2, \ldots, \vecX_\np)$.  In a supervised-learning task, the loss function is designed so that when minimised, each output vector $\vecY_i=f(\vecX_i, \vecW)$ matches as possible closely some given data label or desired value $\vec{y}^*_i$, for each input vector $\vecX_i$ for $i \in \{1,\ldots,\np\}$.  In unsupervised tasks, the loss function would represent some other objective, for example a penalty in a reinforcement-leaning problem, or an ability to reconstruct or group the input data.

The loss function measures how well the NN is achieving its desired task and its value at each point in weight space creates a surface, which the training process attempts to traverse to find a suitably low point.  Most training algorithms use the gradient of the error function with respect to the weights, $\dEdw$, which is calculated by the celebrated backpropagation algorithm \citep{werbos1974, RumelhartHintonWilliams1986}.   Two major difficulties for training are that the loss surface can be very crinkly in places, making the algorithms very slow, and also that the surface may be riddled with sub-optimal local minima and saddle points. 
It is these problems that the various training algorithms in existence, including novel activation functions and weight-initialisation schemes, are designed to overcome, to varying extents.

When a NN processes an input vector $\vecX$, as illustrated in Figure \ref{fig:ffnnSchematics}, the internal (hidden) neurons and output neurons in it will fire at different strengths, or {\it activations}.  Hence there is a real number, the activation strength, associated with each node.  These activation values can be gathered together for all hidden layers and the output layer to form a single vector, $\vecA$. 

Hence for each input vector $\vecX_i$, and given set of weights $\vecW$, there will be an associated activation vector $\vecA_i$.  Given the NN weights $\vecW$ and several input vectors $\vecAllX$, the set of vectors $\vecAllA$ is uniquely determined by the equations that govern the NN's operation.  Conversely, given an arbitrary set of \textit{target} activation vectors, $\vecAllA$, and corresponding input vectors, $\vecAllX$, a relatively cheap calculation using linear algebra could take place to uniquely determine the weight vector $\vecW$ that most closely achieves the set of target-activation vectors.
Therefore the training process could work by iteratively improving the targets, instead of the weights.  That is the central idea of this paper: to do NN training in target space (the space of all possible sets $\vecAllA$) instead of the usual weight space (the space of all possible $\vecW$).

The motivation for switching from weight-space learning to target space is now discussed.    With weight-space learning, any small adjustment to a weight in an early layer shown in Fig. \ref{fig:ffnnSchematics} will make the activations coming out of that layer change by a correspondingly small amount.  However these changed activations will have a knock-on effect in changing the activations in the next layer, and so on with each subsequent layer, often forming a {\it cascade} of changes which reverberate through the later layers.  

If the subsequent layers' neurons are all close to their firing thresholds, or are on a particularly steep part of the activation function, then the small change in the early layer could have a {\it catastrophic scrambling} effect on the NN output.  This is why the error surface in weight space is so crinkly, or even chaotic \citep{skorokhodov2019loss,phan2013error}.  This is not a desirable property for any learning strategy to have to cope with.  Another way of stating that a small change to a weight causes a catastrophic scrambling of behaviour, is to say that the sensitivity of the loss function with respect to that weight is very large.   This is referred to as the exploding-gradients problem \citep{hochreiter1997long}, and we hypothesise that this is the main reason why NNs with many layers are difficult to train using standard backpropagation.

With target space, any small change to the targets for one layer will still cause a correspondingly small change to the activations of that layer.  But then the algorithm that tries to match the node activations to their targets in the subsequent layers will try to choose the weights intelligently so the disturbance to later layers is minimised, an effect which we call {\it cascade untangling} (see Fig. \ref{fig:bagatelleAnalogy}).  If successful, this should minimise the disturbance caused by the initial small change, and hence make the error surface in target space much smoother than that of weight space, directly addressing the exploding-gradients problem.  Increased smoothness of the surface will also reduce the number of local minima in it, and make the crevices in it wider and easier to follow by gradient descent.  
This should be increasingly beneficial for NNs with many layers, and even more so for recurrent neural networks (RNNs) where the output of a neural network is looped back to be combined with subsequent inputs, causing data to cycle around the network many times.   We discuss target-space techniques for RNNs in Section \ref{sec:rnnsTargetSpace}, but initally focus on feed-forward networks. 

If the cascade untangling of target-space learning works as intended, then the path explored by gradient-descent should be more direct and hence reach lower minima.  This could contribute to better generalisation by the neural network \citep{nakkiran2020deep}.  Furthermore, the resulting loss-function surface in target space should be smoother in general, and in particular it may be flatter at the final resting place of the optimisation process.  This could also contribute to better generalisation, since flat minima are hypothesised by \cite{hochreiter1997flat} to produce better generalisation than a sharper minimum (although this is an area for further research because it  might not be straightforward to directly compare the flatness between two different parameterisations of a loss-surface \citep{dinh2017sharp}).

The experimental results given in this paper show that using target space does indeed allow for gaining better performance in the training of deeper networks than occurs with weight space, and includes examples of improved generalisation and improved number of training iterations required for feed-forward networks, recurrent networks and convolutional layered networks; but with a higher computational cost per training iteration (due to the linear algebra process which converts from target space to weight space).  We argue that this extra cost motivates choosing deeper but narrow network architectures, when training a network in target space.
\def\layersep{20}
\usetikzlibrary{arrows.meta}
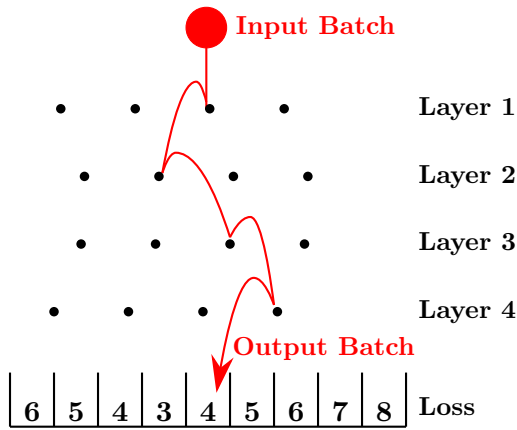
\begin{figure}[ht]
\begin{center} 
\begin{tikzpicture}
    [ x=.5mm, y=.5mm,
      font={\bfseries},
      pins/.style={fill=black},
      scale=0.9
    ]

    \draw[red,fill=red] (9,-22+\layersep*3+4) circle (6); 
    \draw [red, thick]  (9,-22+\layersep*3+4) -- (9,-42+\layersep*3);
    \draw[black]  node[red,midway,right,shift={(15, -22+\layersep*3+4)}] {Input Batch};

	\draw[red, thick]  (9,-40+\layersep*3) parabola bend (6,-40+\layersep*3+6) (-4,-40+\layersep*2-1);
    \draw[red, thick]  (-4,-40+\layersep*2-1) parabola bend (0,-40+\layersep*2-1+6) (16,-40+\layersep);
    \draw[red, thick]  (16,-40+\layersep) parabola bend (22,-40+\layersep+6) (29,-40);
    \draw[red,-{Stealth[length=5mm, width=3mm]}, thick]  (29,-40) parabola bend (23,-32) (12,-66);
    \foreach \i in {-1.5, -0.5, ..., 1.5} {
        \path[pins, shift={(\i * 22, ((3 * \layersep)))}](-1, -42) circle [radius=1.4];
        \path[pins, shift={(\i * 22, ((2 * \layersep)))}](6, -42) circle [radius=1.4];
        \path[pins, shift={(\i * 22, ((1 * \layersep)))}](5, -42) circle [radius=1.4];
        \path[pins, shift={(\i * 22, ((0 * \layersep)))}](-3, -42) circle [radius=1.4];
    }

    \draw [black, thick]  (-49,-76) -- (68,-76);
    \foreach \i in {0, 1, ..., 9} {
        \draw [black, thick, shift={(\i * 13, 0)}]  (-49,-76) -- (-49,-60); 
    }
    \foreach \i in {3, 4, ..., 8} {
        \draw [black, thick, shift={(\i * 13, 0)}]  (-48,-59) node[right,below,shift={(5,-6)}] { \i };
    }
    \foreach \i in {6, 5, ..., 4} {
        \draw [black, thick, shift={(\i * -13, 0)}]  (30,-59) node[right,below,shift={(5,-6)}] { \i };
    }
    \draw[black]  node[midway,right,shift={(69, -42+\layersep*3)}] {Layer 1};
    \draw[black]  node[midway,right,shift={(69, -42+\layersep*2)}] {Layer 2};
    \draw[black]  node[midway,right,shift={(69, -42+\layersep)}] {Layer 3};
    \draw[black]  node[midway,right,shift={(69, -42)}] {Layer 4};
    \draw[black]  node[red,midway,right,shift={(14, -53)}] {Output Batch};
    \draw[black]  node[midway,right,shift={(69, -70)}] {Loss};
    
\end{tikzpicture}
\end{center}      
\caption {In this analogy, a ball bounces deterministically down through a grid of pins, like in the game bagatelle. This represents a neural network processing a batch of input vectors and producing a batch of output vectors. The objective of training the neural network is to arrange the pins to make the ball bounce into a region of minimum loss at the bottom.  Each row of pins represents a layer of weights in the neural network (but the number of pins in each row is unrelated to the number of nodes in that network layer).  The $x$-coordinate of the ball's launch position represents a whole training batch of input vectors, compressed down to a single $x$-coordinate for this visualisation.  Likewise, the ball's horizontal position at each layer of pins represents all of the hidden-state activation vectors at that layer, for each training pattern, compressed down to a single $x$-coordinate.  With weight-space learning, we consider what effect sliding Layer 1 of pins sideways will have on the final destination of the ball -- clearly it will often catastrophically scramble the ball's trajectory (exploding gradients). In contrast, with target-space learning, whenever Layer 1's pins are moved, the positions of the lower rows of pins automatically adjust themselves to try to stabilise the ball's trajectory as much as possible. This represents ``cascade untangling''.  In target space, learning takes place by actively bending segments of the ball's zig-zag trajectory, while causing only minimal disturbances to the other trajectory segments.}  
    \label{fig:bagatelleAnalogy}
\end{figure}
The rest of the paper is structured as follows. In the rest of this section, we discuss related published work.  In Section \ref{sec:algorithm} we define the main target-space algorithm for feed-forward layered neural networks, and then discuss background technical information about the method in Section \ref{sec:technicalInformation}.  In Section \ref{sec:deepArchitectures} we show how the method can be extended to convolutional and recurrent neural networks.  In Section \ref{sec:experiments}, we give experimental results for feed-forward, convolutional and recurrent neural networks.  Finally, in Section \ref{sec:conclusions}, we give conclusions.

\subsection{Related work}

Target-space techniques were originally proposed by \cite{rohwer90} under the name of ``moving targets'', and then re-proposed under different names by \cite{atiya00new,castillo06}. There are some technical difficulties with these early works, which were later identified and improved upon by  \cite{carreira2014distributed}, and follow-up work.  These prior target-space methods, and their modern variants, are described in more detail in Section \ref{sec:previousTargetSpaceResearch}.

Other modern deep-learning methods enable the training of deep networks in a different way from target space.  Some of these are described here.
 
Exploding gradients in deep neural networks were first analysed by \cite{bengio1994learning} and \cite{hochreiter1997long}.  They also identified and defined the opposite problem, vanishing gradients, which also occurs in deep and recurrent networks.  The solution proposed by  \cite{hochreiter1997long}, Long Short-Term Memory (LSTM) networks, focuses on solving vanishing gradients in recurrent networks, and is very effective, especially at spotting and exploiting patterns in long time sequences.  The target-space solution we propose focuses only on addressing exploding gradients, but when combined with a powerful optimiser like Adam, can also learn and exploit long time sequences (even compared to LSTM networks); as shown in Sections \ref{sec:rnnExperiments}-\ref{sec:imdb_rnn}.

\cite{glorot2011deep} identified that vanishing and exploding gradients could largely be controlled by changing the non-linear functions used which  affect the node's firing activation.  They proposed to replace the conventional logistic-sigmoid and hyperbolic-tangent function by a rectified linear function, $\relu(x)$. Since their proposed activation function has a maximum gradient of 1, it limits the scale of a cascade of changes arising from any perturbed weight, and hence eases training of deep networks.  It does not entirely prevent the gradients from decaying/exploding though, since the magnitude of the gradients are also amplified proportional to the magnitude of the weights in each layer \citep{hochreiter1997long}.  Furthermore, the rectified linear function produces some problems of its own, with its unbound magnitude of its output; which can lead to infinities appearing, particularly in recurrent networks.  These infinities make the proposed $\relu$ activation function inappropriate for recurrent networks.  We compare and include our method with a variant of this activation function in Section \ref{sec:experiments}.

Another significant recent breakthrough in training deep networks has been through the careful choice of the magnitude by which weights are randomised before training commences.  The magnitudes derived by \cite{glorot2010understanding} and \cite{he2015delving} are carefully chosen so that the mean and variance in activations of each node remain 0 and 1 respectively, regardless of the depth of the network. This prevents the activations at each layer growing without bound, or saturating on the flat parts of the $\tanh$ activation function, and thus prevent gradients from decaying or exploding.

Batch Normalisation (BN) \citep{ioffebatchnorm2015} is a powerful method for helping with the training of deep networks.  This method can be viewed as a simplification and close relative of target space, and also similar in aim as the above weight-initilisation methods, in that BN prevents the activations of nodes at subsequent layers from growing or saturating without bound. BN works by setting an individual ``target'' for the mean $\mu$ and standard-deviation $\sigma$ for every node in a layer.  These are applied to normalise the entire training batch passing through the given node.  This normalisation can help by performing some limited form of cascade untangling, but to a lesser extent than target space does,  since with BN the targets are just summary statistics for a whole node.  BN is proven to work well in practice, and there has been some discussion on how it works so well \citep{santurkar2018does}.  BN also has a relatively low computational cost compared to target space.  However target space can do a better job of cascade untangling and training deep networks.  We describe empirical comparisons of BN to target space in Section \ref{sec:experiments}. 

\section{Target-Space Algorithm for Layered Feed-Forward Networks} \label{sec:algorithm}

In the first two subsections we describe the notation for ordinary weight-space learning for neural networks.  The target-space algorithm is then defined in the subsequent subsections.

\subsection{Terminology, feed-forward and training mechanisms for a Neural Network} \label{sec:NNDefinition}

We extend the basic NN architecture described in Figure \ref{fig:ffnnSchematics} to act on a batch of size $\nb$ patterns simultaneously.  Concatenate the batch of input column vectors $\{\vecX_{b_1},\vecX_{b_2},\ldots,\vecX_{b_\nb}\}$ side by side into a single matrix $X$ with $\nb$ columns.  Then we can define a feed-forward neural network (FFNN) as a function that maps this matrix, $X$, to an output  matrix, $Y$.  The network is split into $\nl$ layers of nodes, each node having an activation function, $g:\Reals \rightarrow \Reals$, and there being a matrix of weights between each pair of layers.  The activation function $g$ is usually smooth, monotonic and non-linear.  Common choices are $g(x)=\tanh(x)$ or the $\relu$ function \citep{glorot2011deep}.

The layers, respectively, consist of $d_1$, $d_2$, $\ldots$, $d_{\nl}$ nodes, as shown in Figure \ref{fig:ffnnSchematics}.  Thus $X \in \Reals^{d_1 \times \nb}$ and $Y \in \Reals^{d_\nl \times \nb}$.  
\newcommand{\feedinto}[1]{\mathbb{I}\left({#1}\right)} 
In the most general case, each layer $j$ is connected to each later layer $k>j$, via a matrix of weights $W_{j,k} \in \Reals ^{d_k \times d_j}$.  The network is then said to have ``all shortcut connections''.  However in the more common case, shortcut connections are not included and the only non-zero weight matrices are between consecutive layers. 

Each node has a bias which can be implemented by having an extra ``layer 0'' which contains just one node that always has activation of unity.  Thus for each layer $j$, $W_{0,j} \in \Reals ^{d_j \times 1}$ is a column vector of weights coming from layer 0, which represent bias values for layer $j$.

The activations are calculated layer-by-layer, according to Algorithm \ref{alg:feed-forwardDynamics}.   We allow the function $g$ to be applied to a vector or matrix in an elementwise manner, i.e. $(g(A))^{ij}\isDefinedToBe g(A^{ij})$, for all $i$ and $j$.  In line \ref{line:ffdnSums} of the algorithm, $\feedinto{j}$ denotes the set of integer layer-numbers of all layers that {\it feed forwards into} layer $j$.  So for example, for a fully-connected layered network with all shortcut connections, $\feedinto{3}=\{0,1,2\}$.

\begin{algorithm}[h]
\caption{Feed-Forward Dynamics}
\label{alg:feed-forwardDynamics}
\begin{algorithmic}[1]
\STATE $A_0 \leftarrow [1\ 1\ \ldots \ 1]$ \COMMENT{Bias nodes $ \in \Reals^{1 \times \nb}$; a row vector of 1s}
\STATE $A_1 \leftarrow X $ \COMMENT{Input matrix.  $X\in \Reals^{d_1\times \nb}$.}
\FOR{$j=2$  to $\nl$}
\STATE $S_j \leftarrow \sum_{k \in \feedinto{j}}W_{k,j} A_k$ \COMMENT{Sums received by each node. $S_j \in \Reals^{d_j\times \nb}$.}\label{line:ffdnSums}
\STATE $A_j \leftarrow g(S_j)$ \COMMENT{Apply activation function.  $A_j \in \Reals^{d_j\times \nb}.$}
\ENDFOR
\STATE $Y \leftarrow A_{\nl}$  \COMMENT{Output Matrix.  $Y\in \Reals^{d_{\nl}\times \nb}$.}
\end{algorithmic} 
\end{algorithm} 

Running the feed-forward algorithm with an input matrix $X$ generates a sequence of intermediate work-space matrices, $A_j$ and $S_j$ for all layers $j$, whose elements hold the activations and sums, respectively, of each layer's nodes.  These matrices and the output matrix $Y$ are to be retained for later use.  The $p^{\mathrm{th}}$ column of each matrix $X$, $A_j$, $S_j$ and $Y$ all correspond to the same pattern $p$.

To train the neural-network, we first define the {loss function}, or {error function},  $\errorFunction:(X,\vecW)\rightarrow \Reals$, where $\vecW$ is a vector of all of the weights in the network.  For supervised learning, the most common loss functions are the mean-squared error and cross-entropy loss.  Then, we seek to minimise $\errorFunction$ with respect to $\vecW$ using gradient descent:
\begin{align}
 \Delta \vecW = -\eta \dEdw. \label{eqn:weightSpaceGradientDescent}
\end{align}
This weight update is applied iteratively, with a small positive learning rate $\eta$.  The learning rate $\eta$ can be changed over training time, or a more advanced optimiser could be used to try to accelerate learning (e.g. RPROP \citep{riedmiller93direct}, conjugate gradients \citep{moller93scaled}, Levenberg-Marquardt \citep{bishop95}, RMSProp \citep{tieleman2012lecture}, or Adam \citep{kingma2014adam}).

To compute the gradients in the right-hand side of \eqref{eqn:weightSpaceGradientDescent} efficiently, we can use the back-propagation algorithm \citep{werbos1974,RumelhartHintonWilliams1986}, or equivalently automatic differentiation packages provided by a neural-network software library \citep{autodiff,Werbos2005BDi,tensorflow}. 

\subsection{Stacked Layer Input-Matrix and Weight-Matrix Notation}

For layer $j$, define $\layerInputStack{j}$ to be shorthand form for a vertically stacked block matrix of all the $A_k$ matrices that provide an input to layer $j$, i.e. for all the $k \in \feedinto{j}$.  For example, for a simple layered feed-forward network we would have,
\begin{subequations}\label{eqn:BmatrixDefinition}
\begin{align}
\layerInputStack{j} \isDefinedToBe \begin{pmatrix}A_0 \cr A_{j-1}\end{pmatrix},\label{eqn:AmatrixWithBiasCombined} \end{align}
(where $A_0$ is the layer of bias nodes), and if all shortcut connections were present, then this would become, 
\begin{align}
\layerInputStack{j} &\isDefinedToBe \begin{pmatrix}A_0 \cr A_1 \cr \vdots \cr A_{j-1}\end{pmatrix}. \end{align}
\end{subequations}

Also define $\layerWeightStack{j}$ as a side-by-side block concatenation of all the weight matrices that input to layer $j$.  For example, with for a simple layered feed-forward network, we would get:
\begin{subequations}\label{eqn:StackedWmatrixDefinition}
\begin{align}
\layerWeightStack{j} \isDefinedToBe \begin{pmatrix}W_{0,j} & W_{(j-1),j}\end{pmatrix},\label{eqn:WmatrixWithBiasCombined}
\end{align}and, if all shortcut connections were present, we would get,
\begin{align}
\layerWeightStack{j} &\isDefinedToBe \begin{pmatrix}W_{0,j} & W_{1,j} & \ldots & W_{(j-1),j}\end{pmatrix}.\end{align}
\end{subequations}

This simplifies the formula for the NN feed-forward equations;  line \ref{line:ffdnSums} of Algorithm \ref{alg:feed-forwardDynamics} becomes, 
\begin{equation}
S_j \leftarrow \layerWeightStack{j}\layerInputStack{j}. \label{eqn:S_ffnnEconomicalNotation}
\end{equation}

\subsection{Using Targets to Parameterise a Neural Network Instead of Weights} \label{sec:layeredRealisationMain}

So far the neural-network parameters have been the weights $\vecW$.  We now describe how we can switch the representation to ``targets''.  

Define the matrices $T_2$, $T_3$, \ldots, $T_\nl$, to be the ``target matrices'' for each layer. These have the same dimensions as the corresponding $S_j$ matrices.  In the target-space approach, the set of $T_j$ matrices will be the learnable parameters, replacing the role of the weight matrices.  The weight matrices are relegated into calculated quantities that are dependent on the $T_j$ matrices.

The target matrix for each layer $T_j$ holds the ``targets'' for the $S_j$ matrix at that layer; hence we want to choose the weights which make the $S_j$ matrices get as close as possible to the $T_j$ matrices, or to minimise $\norm{S_j-T_j}$, where $\norm{\cdot}$ denotes the Frobenius norm.  To simplify computational complexity, we do this in a greedy layer-by-layer manner.

Substituting \eqref{eqn:S_ffnnEconomicalNotation} shows that we therefore need to find 
\begin{align}
\layerWeightStack{j}=\arg \min_W \left[\norm{W\layerInputStack{j}-T_j}^2+\lambda \norm{W}^2\right], \label{eqn:leastSquaresCost} 
\end{align}
where the $\lambda \norm{W}^2$ term is included to provide Tikhonov regularisation, which ensures that the solution in $W$ is unique and kept reasonably small.  The minimisation in \eqref{eqn:leastSquaresCost} is a standard least-squares problem from linear algebra, with solution
\begin{equation}\layerWeightStack{j} = T_j{\left(\layerInputStack{j}\right)^\pseudoinv}, \label{eqn:WlayeredDiscreteFCCRealisation}
\end{equation}
where the $\pseudoinv$ indicates a regularised pseudoinverse matrix, defined by
\begin{equation}
A^\pseudoinv \isDefinedToBe {A}^T( A{A}^T+\lambda I)^{-1}. \label{eqn:BdaggerDefinition}
\end{equation}

Here $A{A}^T$ is referred to as the Gramian matrix, $\lambda\geq0$ specifies the amount of Tikhonov regularisation, and $I$ is the identity matrix.  The presence of $\lambda I$ in \eqref{eqn:BdaggerDefinition} prevents the occurrence of non-invertible matrices.\footnote{An alternative to Tikhonov regularisation would be to use the Truncated Singlular Value Decomposition pseudoinverse, but this was avoided because the truncation means the derivatives are not as smooth. However the SVD (or similar decompositions) may be used  to implement \eqref{eqn:BdaggerDefinition} in practice, to obtain improved numerical stability.}

Hence the layer weights and activations can be calculated layer by layer.  The full method by which the weights are calculated from the target matrices is given in Algorithm \ref{alg:mappingTargetsToWeightsSCU}. 

\begin{algorithm}[h]
\caption{Converting Targets to Weights, in a FFNN, with Sequential Cascade Untangling (SCU)}
\label{alg:mappingTargetsToWeightsSCU}
\begin{algorithmic}[1]
\STATE $\ovA_0 \leftarrow [1\ 1\ \ldots \ 1]$ \COMMENT{Bias nodes. $\ovA_0 \in \Reals^{1\times \nbt}$.}
\STATE $\ovA_1 \leftarrow \overline{X}$ \COMMENT{Input matrix. $\overline{X} \in \Reals^{d_1\times\nbt}$.}
\FOR{$j=2$  to $\nl$}
\STATE $\layerWeightStack{j} \leftarrow T_j{\left(\layerInputStackOv{j}\right)^\pseudoinv}$ \label{line:ldfr:Wequals} \COMMENT{Calculates weights to layer $j$.  $T_j \in \Reals^{d_j\times \nbt}$.}
\STATE $\ovS_j \leftarrow \layerWeightStack{j} \layerInputStackOv{j}$ \COMMENT{$\ovS_j \in \Reals^{d_j\times \nbt}$.}\label{line:ldfr:Sequals}
\STATE $\ovA_j \leftarrow g(\ovS_j)$ \label{line:ldfr:Aequals} \COMMENT{$\ovA_j \in \Reals^{d_j\times \nbt}$.}
\ENDFOR
\end{algorithmic} 
\end{algorithm} 

The main inputs to this algorithm are an input matrix $\overline{X}$ with batch size $\nbt$, and a list of target matrices $T_j$.  The main outputs of this algorithm are the realised weight matrices, $W_j$.  The quantities $\ovS_j$ and $\ovA_j$ are work-space matrices.

Because $\layerInputStackOv{j}$ is a shorthand for a stack of activation matrices $\ovA_j$, as defined in \eqref{eqn:BmatrixDefinition}, it is intended that the changes to $\ovA_j$ in line \ref{line:ldfr:Aequals} will immediately affect the $\layerInputStackOv{j}$ matrices referenced in line  \ref{line:ldfr:Wequals} for higher values of $j$.   This is what carries forwards the changes of an earlier layer, so that they can be corrected for by a later layer.

Once these weight matrices are obtained, they are then used in Alg. \ref{alg:feed-forwardDynamics} to calculate the actual NN output.  Note that Alg. \ref{alg:mappingTargetsToWeightsSCU} followed by Alg. \ref{alg:feed-forwardDynamics} are run back-to-back, in that order, and can therefore be viewed as one continuous computational graph.   \citep[This is in contrast with some prior published work on target space, e.g.][where there are alternating phases of updating the $W_j$ matrices followed by updating $T_j$ matrices.  In our method, the $W_j$ matrices are defined as functions of the $T_j$ matrices, and there are no alternating phases.]{carreira2014distributed}

Alg. \ref{alg:mappingTargetsToWeightsSCU} is designed to work with a potentially different input batch $\overline{X}$ from the input matrix $X$ used to evaluate the output of the main network via Alg. \ref{alg:feed-forwardDynamics}.  This separation aids using mini-batches when training the network, which is discussed further in Sec. \ref{sec:minibatching}. Note that because Alg. \ref{alg:mappingTargetsToWeightsSCU} uses a different input matrix ($\overline{X}$) compared the input matrix $X$ used by  Alg. \ref{alg:feed-forwardDynamics}, therefore the work-space matrices $S_j$ and $A_j$ are a different set of variables in Alg.  \ref{alg:mappingTargetsToWeightsSCU}, compared to Alg. \ref{alg:feed-forwardDynamics}.

Note that the aim of matching the $\ovS_j$ matrices to their targets by \eqref{eqn:leastSquaresCost} will not be achieved exactly.  In general where the number of patterns $\nbt$ is larger than the rank of the weight matrix, matching the targets exactly will be impossible.  Hence we carry forward the disturbances actually achieved to the $\ovS_j$ matrices, as opposed to the disturbances intended by $T_j$ matrices, in Line \ref{line:ldfr:Aequals} of Alg. \ref{alg:mappingTargetsToWeightsSCU}. Then the subsequent layers' targets will act to continue to try to dampen down this disturbance, taking into account the fact that the previous layer's targets will not have been met exactly, so that the subsequent cascade of changes is always minimised as much as possible.  Hence we refer to the algorithm as having {\it Sequential cascade untangling} (SCU). 

We found SCU to be much more effective when training neural networks than an alternative of assuming targets are met exactly, which would be implemented by replacing line \ref{line:ldfr:Aequals} of Alg. \ref{alg:mappingTargetsToWeightsSCU} by 
\begin{align}
\ovA_j \leftarrow g(T_j).\label{eqn:mccApproximation}
\end{align} Since this approach does not carry forwards the actual cascade of changes beyond just one layer, we call this alternative approach ``optimistic  cascade untangling'' (OCU), and this is what prior published research \citep[for example,][]{rohwer90,atiya00new,castillo06} has always done.  Experiments in Sec. \ref{sec:twoSpiralsExperiments} (Fig. \ref{fig:resultsTwoSpiralsFCC_MCC_bn}) show a significant improvement in performance from using SCU over OCU on the Two-Spirals classification problem, and experiments in Sections \ref{sec:rnnExperiments} and \ref{sec:imdb_rnn} show the advantage it gives in recurrent neural networks.

\subsection{Calculating the Learning Gradient in Target Space} \label{sec:dEdTcalculation}
The previous subsection described an algorithm which converts targets to weights.   The next objective is to be able to do gradient descent in target space, i.e. with respect to the targets themselves.  

Algorithm \ref{alg:mappingTargetsToWeightsSCU} can be viewed as a mapping function $\mappingFunction$ from targets to weights, such that \begin{align}\vecW=\mappingFunctionFull\label{eqn:mappingFunctionDefinition},\end{align}where $\vecT$ is a shorthand for the vector of all target matrices flattened and concatenated together.  Given such a differentiable mapping function, $\mappingFunction$, we can define the loss function $L$ in terms of the targets (which we will denote as $\errorFunctionT$), as follows:
\begin{align}
\errorFunctionT(X,\vecT) \isDefinedToBe \errorFunction(X,\mappingFunctionFull) \label{eqn:errorFunctionTDefinition}
\end{align}

Consequently, using the chain rule we can convert gradient descent in weight space to gradient descent in target space:
\begin{align}
    \dEdt=\left(\fracpartial{\mappingFunction}{\vecT}\right)^T\dEdw, \label{eqn:connectionWSGradientToTSGradient}
\end{align}
where $\fracpartial{\mappingFunction}{\vecT}$ uses Jacobian matrix notation, and $\dEdw$ and $\dEdt$ are treated as column vectors.

This gradient $\dEdt$ allows us to perform gradient descent in target space, directly on the main neural-network objective function, via
\begin{align}
 \Delta \vecT = -\eta \dEdt. \label{eqn:targetSpaceGradientDescentStep}
\end{align}

Algorithm \ref{alg:layeredDiscreteFCCRealisationDEdT} applies \eqref{eqn:connectionWSGradientToTSGradient} to calculate the $\fracpartial{\errorFunctionT}{T_j}$ matrices, for Algorithm \ref{alg:mappingTargetsToWeightsSCU}'s mapping method. In this code, $\hadamard{A}{B}$ means the Hadamard or elementwise product. The algorithm uses workspace matrices $\delta \ovA_j$ and $\delta \ovS_j$, which are identically dimensioned to their non-prefixed counterparts, for each layer $j$.  The matrix $\delta \layerInputStackOv{j}$ is built up of $\delta \ovA_k$ matrices, in the same way as Equation \eqref{eqn:BmatrixDefinition}, and similarly $\fracpartial{\errorFunction}{\layerWeightStack{j}}$ is composed of $\fracpartial{\errorFunction}{W_{k,j}}$ matrices like \eqref{eqn:StackedWmatrixDefinition}.  It is assumed that these matrices point to the same underlying data, so for example, changing $\delta \layerInputStackOv{3}$ will immediately affect $\delta \ovA_2$, and vice versa.

\begin{algorithm}[h]
\caption{Calculation of Learning Gradient in Target Space} 
\label{alg:layeredDiscreteFCCRealisationDEdT}
\begin{algorithmic}[1]
\REQUIRE $\ovS_j$, $\ovA_j$ and $W_j$ matrices calculated by Alg.
\ref{alg:mappingTargetsToWeightsSCU} for input matrix $\overline{X}$, and $\fracpartial{\errorFunction}{W_{k,j}}$  matrices calculated by back-propagation applied to Alg. \ref{alg:feed-forwardDynamics} for input matrix $X$. 
\STATE $\forall j, \delta \ovA_j \leftarrow 0$
\FOR{$j=\nl$ to $2$ step $-1$}
\STATE $\delta \ovS_j \leftarrow \hadamard{(\delta \ovA_j)}{g'(\ovS_j)} $ \label{line:ldfrdedt:deltaSequals}
\STATE $\fracpartial{\errorFunctionT}{T_j} \leftarrow \left (\fracpartial{\errorFunction}{\layerWeightStack{j}} + (\delta \ovS_j){\layerInputStackOv{j}}^T\right)(\layerInputStackOv{j}^\pseudoinv)^T $ \label{line:ldfrdedt:dEdTequals}
\STATE $\delta \layerInputStackOv{j}\leftarrow \delta \layerInputStackOv{j}+{\layerWeightStack{j}}^T(\delta \ovS_j-\fracpartial{\errorFunctionT}{T_j}) + (\layerInputStackOv{j} \layerInputStackOv{j}^T+\lambda I)^{-1} \left(\left(\fracpartial{\errorFunction}{\layerWeightStack{j}}\right)^T + {\layerInputStackOv{j}}(\delta \ovS_j)^T \right)(T_j-\ovS_j)$
\label{line:ldfrdedt:deltaBequals} 
\ENDFOR
\end{algorithmic} 
\end{algorithm} 

The useful outputs of the algorithm are the quantities $\fracpartial{\errorFunctionT}{T_j}$, for all layers $j$,  which can be written collectively as $\dEdt$.  Hence the algorithm gives $\dEdt$, which can be used to perform gradient descent in target space \eqref{eqn:targetSpaceGradientDescentStep}. As with weight-space gradient descent, a more advanced optimiser might be applied to achieve a speed up.  

The target-gradient computation algorithm (Alg. \ref{alg:layeredDiscreteFCCRealisationDEdT}) is derived in Appendix \ref{sec:alg:derivLayeredDiscreteFCCRealisationDEdT}.  
The most interesting part of the derivation is the differentiation under the matrix inverse operation.    This was omitted by prior research \citep{rohwer90,castillo06}, which indicates that their learning gradients were incorrect. Our informal experiments (not recorded here) showed that this severely reduced performance of those prior algorithms.  Modern automatic-differentiation  \citep{tensorflow} libraries correctly handle differentiation under a matrix inverse, but as this step is non-obvious to derive manually, we have included the explicit algorithm here.  Alternatively, if Alg. \ref{alg:mappingTargetsToWeightsSCU}  followed by Alg. \ref{alg:feed-forwardDynamics} followed by the calculation of $\errorFunction$ is passed through an automatic-differentiation library, then $\dEdt$ will be calculated correctly, automatically. 

The algorithmic complexity to implement one iteration of target-space learning is derived (under various assumptions) in Appendix \ref{sec:appendixComputationalComplexityFFNN} to be approximately $4\nbt/\nb$ times larger than time taken to implement one iteration of weight-space learning. Note that in this ratio, $\nbt$ is the batch-size used for the target space matrix $\overline{X}$, and $\nb$ is the batch size for the weight-space input matrix $X$.  Hence if smaller mini-batches are used to acquire the weight-space gradient than are used in the target-space algorithms, then the time per iteration of the target-space algorithm (which cannot use tiny mini-batches) would become increasingly large in comparison to the weight-space calculations.  Hence in the extreme case of pattern-by-pattern learning ($\nb=1$), the target-space algorithm would be slower by a very significant factor of approximately $4\nbt$.  In the experiments of Section \ref{sec:twoSpiralsExperiments}, we use $\nbt=\nb$, and the resulting theoretical ratio of 4 holds out well empirically.

\section{Technical Aspects for Target-Space Implementations} \label{sec:technicalInformation}

The previous section has defined the main target-space method. We now consider some technical aspects, including how to use mini-batching, the effects of choice of $\lambda$, how to initialise the target variables at the start of training, detail of differences between this method and previous published target-space work, and convergence properties of our method.

\subsection{Mini-batching and the Choice of $\overline{X}$} \label{sec:minibatching}

For very large datasets, it becomes prohibitively expensive to compute $\dEdw$ for the whole dataset.  Hence with very large datasets, it is standard practice in deep-learning to use mini-batches; that is to operate on a smaller, randomly chosen, subset of the training data in any one training iteration, with $\nb\ll\np$.  The mini-batch chosen would be used to build the input matrix $X$ inputted to Alg. \ref{alg:feed-forwardDynamics}. Using mini-batching also introduces a stochastic element to the optimisation process, which is also beneficial in finding flatter final minima in the loss-function surface, and thus improving generalisation  \citep{bottou2010large,masters2018revisiting}.

As noted in Section \ref{sec:layeredRealisationMain}, it is possible to use a different $X$ for the computation of $\dEdw$ by backpropagation through Alg. \ref{alg:feed-forwardDynamics} from the $\overline{X}$ used in the target-space calculations of Algs. \ref{alg:mappingTargetsToWeightsSCU}-\ref{alg:layeredDiscreteFCCRealisationDEdT}. But unlike the random mini-batches which may be used for calculating $\dEdw$, the $\overline{X}$ used for target space must be fixed; because every time we shuffle the mini-batches in $\overline{X}$, the corresponding learnable quantities $T_j$ would have their meaning scrambled, which would disrupt learning.

For computational efficiency, it is possible for the patterns in $\overline{X}$ to be a mini-batch, i.e. a subset of the entire training set, or even a fixed random matrix\footnote{See Sec. \ref{sec:imdb_rnn} for an example of this.}. But it must be a \textit{fixed} matrix.   

The larger $\nbt$ is (where $\nbt$ is the number of columns in $\overline{X}$), the more computationally expensive things will become.  So how large should $\nbt$ be? Ideally, $\nbt$ should be sufficiently large so that the Gramian matrix in \eqref{eqn:BdaggerDefinition} would not have any zero eigenvalues.  The more non-zero eigenvalues this product has, i.e. the more linearly independent columns in each $A_j$, the more useful the pseudoinverses calculated will be in performing cascade untangling (defined in Section \ref{sec:introduction}).  If there are too few patterns in $\overline{X}$ then it will mean that target-space learning will not be able to generate usefully full-rank weight matrices in any layer where the number of layer inputs exceeds $\nbt$, which can limit the representation capabilities of the neural network (see section \ref{sec:twoSpiralsExperiments}, Fig. \ref{fig:hyperParameterSensitivity1}, for an example.)

Since the side-dimension of $\ovA_j \ovA_j^T$ is equal to the number of inputs to layer $j$, as a rule of thumb, we recommend to set $\nbt$ to be preferably as large as the widest layer in the network, and more so if the computational expense can be spared; as this will usually ensure the Gramian matrix is full rank. Achieving this while also maintaining computational efficiency motivates the use of network architectures which are deep and narrow, as opposed to architectures with a large number of nodes to each hidden layer.

\subsection{Choice of $\lambda$} \label{sec:choiceoflambda}

For choosing $\lambda$ in equation \eqref{eqn:BdaggerDefinition}: if it is too large then the effect of the pseudoinverses in \eqref{eqn:BdaggerDefinition} will be dulled in their ability to perform cascade untangling.  Hence for large $\lambda$, the benefits of target-space learning start to disappear.

If $\lambda$ is too small, then the inverse might become close-to-singular.  This would mean small changes in $A_j$ make large changes to the generated weight matrices, and hence the learning gradients in target space would become too steep.  

If instability in learning is observed, then $\lambda$ could be increased, to try to remove any particularly steep gradients in target space caused by the matrix inversion process.  We used either $\lambda=0.001$ or $\lambda=0.1$ in all experiments in this paper.  

Note that the $\lambda$ in equation \eqref{eqn:leastSquaresCost} is performing L2 regularisation only on the mapping between targets and weights.  It does not limit the final magnitude of the weights in the neural network, since there is no restriction of the magnitude of $T$ in equation \eqref{eqn:WlayeredDiscreteFCCRealisation}, and there is no cost on the magnitude of $T$ appearing in the main training objective function $L$. Hence, this L2 regularisation should not be confused with a desire to apply L2 regularisation on the weights of the neural network (weight decay), which would have the intention of regularising the neural network into having smaller magnitude weights.  If that was required, then explicit weight decay terms (on the magnitudes of $W$) should be added into $L$.

\subsection{Target initialisation}
At the start of training, the layer target matrices $T_j$ need to be randomised.  We used a truncated normal distribution, with mean 0 and a fixed variance to randomise each element of each $T_j$ matrix.  

Since these initial layer targets have the same fixed variance at every layer, the variance of the magnitudes of the layer activations should be the same at every layer of the initially-randomised network.  This is in contrast to weight-space initialisation, where unless the initial randomised weight magnitudes are chosen very carefully (such as by using the methods proposed by \cite{he2015delving,glorot2010understanding}), then the activations at subsequent layers can grow exponentially, eventually either saturating or becoming zero.

We have empirically found that it may be beneficial to run Alg. \ref{alg:mappingTargetsToWeightsSCU} once immediately after the initial targets are randomised, to compute the weight matrices and $\ovS_j$,  and then to apply  
\begin{equation}
    T_j \leftarrow \ovS_j,\ \forall j, \label{eqn:initialTargetsProjection}
\end{equation}
exactly once before training commences.  This simply projects the newly-randomised targets on to the hypersurface through target space which represents the subset of targets which are exactly achievable.  This step is done in all of the target space experiments presented in this paper.  It remains to be seen how much value this step adds, although our informal experiments seemed to show some benefit in our recurrent neural-network experiments.

\subsection{Relationship to Prior Target-Space Research} \label{sec:previousTargetSpaceResearch}

The work by \cite{rohwer90} is a stand-out early work on target space which we discuss here, along with more recent notable work, particularly those following on from \cite{carreira2012distributed}.  Some of the prior work is dedicated to recurrent networks \citep[e.g.][]{atiya00new}, some is  dedicated to feed-forward networks with one hidden layer \citep{castillo06}, and some (especially more recent publications) is dedicated to general deep architectures \citep[e.g.][]{rohwer90,carreira2012distributed,lee2015deeply,lee2015difference,taylor2016training,zhang2016efficient,frerix2017proximal}.

In some of the prior works, the process which converts targets into weights seeks to minimise $\norm{g(S_j)-T_j}$ or $\norm{S_j-g^{-1}(T_j)}$ instead of $\norm{S_j-T_j}$.  Unfortunately there is no closed-form solution to minimise $\norm{g(S_j)-T_j}$ with respect to the weights, and the second option $\norm{S_j-g^{-1}(T_j)}$ requires the function $g$ to be invertible and the domain of $T_j$ to be restricted to the range of $g$.

Early prior published work \citep{rohwer90,atiya00new,castillo06} is only applicable to the sum-of-squared loss function, and hence only to supervised regression problems.  A significant defect of these early target-space methods, which probably held back their greater adoption, is that instead of optimising the main objective function $\errorFunction$, they instead optimise an intermediate loss function, similar in concept (ignoring bias and shortcut connections) to
\begin{equation}E(X,\vecT)=\sum_j \norm{W_jg(T_{j-1})-T_j}^2,\label{eqn:errorRohwer}\end{equation}
instead of the true sum-of-squares cost function,
\begin{equation}E(X,\vecT)=\sum_j \norm{W_jg(S_{j-1})-T_j}^2.\label{eqn:errorSumOfSquares}\end{equation} 
They aim to minimise \eqref{eqn:errorRohwer} with respect to the variables $T_j$, subject to each $W_j$ satisfying \eqref{eqn:WlayeredDiscreteFCCRealisation}, and subject to the final layer's targets satisfying $T_\nl=Y^*$, where $Y^*$ is the target data in the supervised regression problem.  If \eqref{eqn:errorRohwer} is successfully minimised down to zero then it will follow that $T_j=S_j$ for all $j$, and \eqref{eqn:errorRohwer} will match \eqref{eqn:errorSumOfSquares}, and so the supervised learning problem will be solved.  However seeing as it is in general impossible to achieve a zero error in \eqref{eqn:errorRohwer}, it means that the first network layer will fail to achieve $S_1=T_1$ exactly, and hence the ``input'' to the second layer in \eqref{eqn:errorRohwer}, namely $g(T_1)$, will be wrong.  This misalignment between $S_j$ and $T_j$ will grow more and more as the layer number $j$ increases.  The end result is that local minima in \eqref{eqn:errorRohwer} do not align with local minima in \eqref{eqn:errorSumOfSquares}, and so gradient descent on \eqref{eqn:errorRohwer} does not actually minimise the intended loss function.   This was a crucial error limiting the applicability of the methods by \cite{rohwer90} and \cite{castillo06}.   Additionally the work by \cite{rohwer90} and \cite{castillo06} make an incorrect derivative calculation in computing the learning gradient, by omitting to differentiate through the matrix inverse operation of equation \eqref{eqn:BdaggerDefinition}.
A related error of following the wrong gradient descent direction appears in the work of \cite{atiya00new}.  They approximate $\fracpartial{\errorFunctionT}{T_j}=0$ for all $j<\nl$, which is incorrect since cascade untangling can never occur perfectly.

Later work rectifies these problems.  The work by \cite{carreira2012distributed} refers to the target variables as auxiliary coordinates.  They solve the problems associated with \eqref{eqn:errorRohwer} by instead using a bespoke objective function that is something like a weighted sum between \eqref{eqn:errorRohwer} and  \eqref{eqn:errorSumOfSquares}, and where the weighting towards \eqref{eqn:errorSumOfSquares} is gradually increased during learning.  This ensures that it is \eqref{eqn:errorSumOfSquares} that is finally optimised, while benefitting from the easier learning of \eqref{eqn:errorRohwer} in earlier training.  However their method requires alternating phases of minimisation with respect to  $W_j$ followed by minimising with respect to $T_j$; and then both of these phases need interlacing with increasing the weighting of \eqref{eqn:errorSumOfSquares} versus \eqref{eqn:errorRohwer}.  Our method streamlines this process by having a single optimisation to do, which avoids zig-zagging through the search space, and allows for acceleration methods to be applied.  But in comparison, their method increases the decoupling of the layers by successfully using an equation based on \eqref{eqn:errorRohwer} for the majority of the learning process. 

\cite{frerix2017proximal} extend upon the work of \cite{carreira2012distributed} but they modify the cost function so that the targets within it are anchored to the forward-propagated activations (by an equation similar to \eqref{eqn:initialTargetsProjection}; so that the targets are no-longer free variables to be learned).  This modification creates an implicit quadratic cost function attached to each layer (similar to \eqref{eqn:errorRohwer}) which enables the use of a semi-implicit optimisation algorithm based on proximal updates.  The proximal updates can converge under much higher learning rates than would be possible with ordinary gradient descent.

In ``Difference Target Propagation'', \cite{lee2015difference} define a method which uses learnable targets for each hidden layer.  In this method, the target at one layer $T_j$ is iteratively set to $L_P^{-1}(T_{j+1})$, where $L_P^{-1}$ is an inverse function of the layer's forward-propagation function, and where this inverse (being generally an unknown function) is learned by an auto-associative network which learns to model $L_P$ for each given network layer. This method potentially allows training of networks with discrete activation functions. In ``Deeply-Supervised Nets'', \cite{lee2015deeply} add an extra support-vector machine classifier for the output of each layer.  This provides extra training information; a kind of target for each hidden layer, which proves very effective in training deep classification networks.

\cite{taylor2016training} use learnable targets for both the $A_j$ and $S_j$ matrices, and update these learnable variables with iterative application of a closed-form Bregman method, which trains the network to solve the objective function, without needing to use any form of gradient descent. \cite{zhang2016efficient} use a similar iterative scheme to train neural networks to generate supervised hash codes.

In summary, much of the prior work shows the potential and power of target space, and the recent prior work addresses the problems appearing earlier in novel ways.  

Our work provides several notable further enhancements and alternatives to the prior work, particularly regarding the introduction of the SCU method, which we show in our experiments is beneficial to performance.  Furthermore, none of the prior work shows how to  separate the input matrix $\overline{X}$ (which is used for calculating the weights from targets) from the input matrix $X$ (which is used to run the neural network in Alg. \ref{alg:feed-forwardDynamics}).    Our work also introduces the correction of gradient calculations through the pseudoinverse operation (which is necessary to apply \eqref{eqn:connectionWSGradientToTSGradient} correctly);  the separation of the main objective's loss function from the intermediate closed-form least-squares minimisation; and the introduction of mini-batches.  The simplicity of the method, and the view of searching in ``target space'', gives a single, simple, gradient-descent objective, i.e. \eqref{eqn:targetSpaceGradientDescentStep}, which can easily be combined with existing acceleration schemes such as Adam.

\subsection{Convergence Properties and Representation Capabilities of Target Space}

The target-space gradient descent update \eqref{eqn:targetSpaceGradientDescentStep} is derived to be true gradient descent on the loss function $\errorFunction(\vecX, \mappingFunctionFull)$ with respect to $\vecT$.  The loss function $L$ is the main learning objective function, as chosen by the practitioner.  For example, for a regression problem, $L$ could be the mean-squared error, or for classification problems, it could be cross-entropy loss. 

A potential source of confusion is that there is a second loss function appearing in the least-squares sub-problem given by \eqref{eqn:leastSquaresCost}, and also that the targets in each layer will not usually be matched exactly; but this least-squares sub-problem is completely separate from the neural-network's main objective function, $L$.  To see this more clearly, the mapping from targets to weights, $\vecW=\mappingFunctionFull$, given by Alg. \ref{alg:mappingTargetsToWeightsSCU}, could be replaced by any other well-defined differentiable mapping function.  Regardless of what the differentiable function $\mappingFunction$ is, and regardless of how well any targets are matched or not matched, gradient descent is still performed on the main neural-network objective function $L$ by \eqref{eqn:connectionWSGradientToTSGradient} and \eqref{eqn:targetSpaceGradientDescentStep}.

Any sufficiently small step size in target space by \eqref{eqn:targetSpaceGradientDescentStep} will yield a decrease in $L$, since, to first order:
\begin{align}
    \Delta \errorFunction &\approx \left(\Delta \vecT\right)^T\dEdt &\text{(ignoring higher order terms)}\nonumber \\
    &= -\eta \left( \dEdt\right)^T \dEdt&\text{(by \eqref{eqn:targetSpaceGradientDescentStep})}\nonumber \\
    &\leq 0 \label{eqn:targetSpaceDefinteDecrease}
\end{align}

Since the function $L$ has a lower bound, $L$ will decrease monotonically but not beyond that bound.  Hence convergence of $\errorFunction(\vecX, \mappingFunctionFull)$ to some limit is guaranteed.  Similarly, the standard convergence proofs for gradient descent with appropriately chosen step sizes apply here \cite[Section 1.2.2]{bertsekas1998nonlinear}.
 
Since the differentiable mapping function $\mappingFunction$ is arbitrary, the convergence guarantees work just as well as for the OCU and SCU variants described in Section \ref{sec:layeredRealisationMain}.  The difference is that we hope that the target-space loss-surface is smoother in one variant than the other (and that both variants are smoother than in weight space), and therefore they will produce faster convergence and better generalisation (which can only be justified empirically; see discussion in Section \ref{sec:introduction} and empirical results in Section \ref{sec:experiments}).

For any given set of weights we can run the neural network forwards, and can capture the sums $\ovS_j$ at each layer $j$, and assign these to the targets at each layer, by \eqref{eqn:initialTargetsProjection}.  Ignoring the Tikhonov regularisation in \eqref{eqn:leastSquaresCost}, this will mean the targets will generate weights approximately equal to the given set of weights that we started with.    This shows that any point in weight space has at least one equivalent representation in target space, such that $\mappingFunction$ is a many-to-one function, and hence any local minimum in weight space could be reached by gradient descent from an appropriate random start point in target space.

An important question is the relationships between ``solutions'' (i.e. stationary points) of the target-space problem, $\errorFunction(\vecX, \mappingFunctionFull)$, and those of the original weight-space one, $\errorFunction(\vecX, \vecW)$. Equation \eqref{eqn:connectionWSGradientToTSGradient} shows that whenever $\dEdw=0$, we must also have $\dEdt=0$.  Furthermore, when Alg. \ref{alg:mappingTargetsToWeightsSCU} is used to define the mapping function $\mappingFunction$, Appendix \ref{sec:appendix:stationaryPoints} shows that whenever $\dEdt=0$, we must also have $\dEdw=0$.  Hence any stationary point in target space is also a stationary point in weight space, and also the reverse is true. 

For a step in target space $\Delta \vecT$, applying a first-order Taylor-Series Expansion of \eqref{eqn:mappingFunctionDefinition} gives:
\begin{align}
    \Delta \vecW &\approx \fracpartial{\mappingFunction}{\vecT} \left(\Delta \vecT\right)&\text{(first-order Taylor Series)}\nonumber \\
    &= -\eta  \fracpartial{\mappingFunction}{\vecT} \left( \dEdt\right)&\text{(by \eqref{eqn:targetSpaceGradientDescentStep})}\nonumber \\
     &= -\eta  \fracpartial{\mappingFunction}{\vecT}\left(  \fracpartial{\mappingFunction}{\vecT}\right)^T \dEdw &\text{(by \eqref{eqn:connectionWSGradientToTSGradient})} \label{eqn:preconditioner} 
\end{align}
Comparing \eqref{eqn:preconditioner} to \eqref{eqn:weightSpaceGradientDescent} shows that a first-order approximation to gradient descent in target space via \eqref{eqn:targetSpaceGradientDescentStep} is equivalent to descent in weight space, but where each weight-space direction is multiplied by a positive semi-definite preconditioner matrix $\fracpartial{\mappingFunction}{\vecT}\left(  \fracpartial{\mappingFunction}{\vecT}\right)^T$.  However by explicitly working in target space, we get the benefit of being able to apply an acceleration procedure to the descent steps in target space, such as Adam, and still retain the convergence guarantees proven for that acceleration method.  We would lose these guarantees if we applied the semi-definite preconditioner matrix in weight space, and then applied Adam afterwards.  Also, rather than viewing target space simply as weight space with this particular preconditioner, we have found empirically that issuing \eqref{eqn:preconditioner} directly can be more unstable than using the exact function $\vecW=\mappingFunctionFull$, presumably due to the first-order approximation used in \eqref{eqn:preconditioner}; although this is an area for further research.

If mini-batching is used to generate samples of $X$, then the expectation of the gradient descent direction in target space can be derived as follows.  Denote the sampled mini-batch as $\hat{X}$, and $\hat{L}\isDefinedToBe \errorFunction(\hat{X},\vecW)$, and $\mathbb{E}_{\hat{X}}$ to be the expectation operator with respect to $\hat{X}$.  Then,
\begin{align}
\mathbb{E}_{\hat{X}}\left(\Delta \vecW \right)&=-\eta \mathbb{E}_{\hat{X}}\left(\fracpartial{\mappingFunction}{\vecT} \fracpartial{\mappingFunction}{\vecT}^T \fracpartial{\hat{L}}{\vecW}\right) & \text{(by \eqref{eqn:preconditioner})}\nonumber\\
&=-\eta \left(\fracpartial{\mappingFunction}{\vecT} \fracpartial{\mappingFunction}{\vecT}^T \right) \mathbb{E}_{\hat{X}}\left( \fracpartial{\hat{L}}{\vecW}\right) & \nonumber\\
&=-\eta \left(\fracpartial{\mappingFunction}{\vecT} \fracpartial{\mappingFunction}{\vecT}^T \right) \fracpartial{L}{\vecW}
.
\end{align}

The second line above follows because the mapping function $\mappingFunction$ is independent of the sample chosen $\hat{X}$, for a given $\vecW=\mappingFunctionFull$.  
The final line concludes that even though mini-batching may be used, with $\overline{X}$ independent of $\hat{X}$, the expectation of the learning gradient in target space will still produce a preconditioned descent step on the loss function on the whole dataset $L$.

\section{Specific Deep Architectures} \label{sec:deepArchitectures} 

The target-space method can be extended to different neural architectures and layer types.  Here we show specifically how the method can be extended to convolutional neural networks  and recurrent neural networks.

\subsection{Application to RNNs} \label{sec:rnnsTargetSpace}

Recurrent neural networks (RNNs) are a powerful architecture of neural networks, which extend the feed-forward network by having one or more recurrent (backward pointing) weights. These feedback connections allow information from previous inputs be retained and to contribute extra information to subsequent inputs to the network.  This creates short-term memory, which allows the network to remember and act on past inputs, enabling a RNN to potentially have much greater functionality than a FFNN, potentially allowing it to act like an agent interacting with an environment.  Successful RNN applications are in areas such as neurocontrol, time-series analysis, image captioning, language translation, and question answering \citep{karpathy2015deep,fairbank14clipping,Fairbank201474,sutskever2014sequence,samothrakis2016ijcnn}.  However RNNs are generally more difficult to train than feed-forward networks, with major challenges being vanishing or exploding learning gradients, making it difficult for a RNN to remember information over long time sequences.   

This section describes how a RNN can be trained in target-space.  Target-space methods potentially allow RNNs to tackle more complex time sequences and data-processing tasks which previously have been very challenging for RNNs to solve.

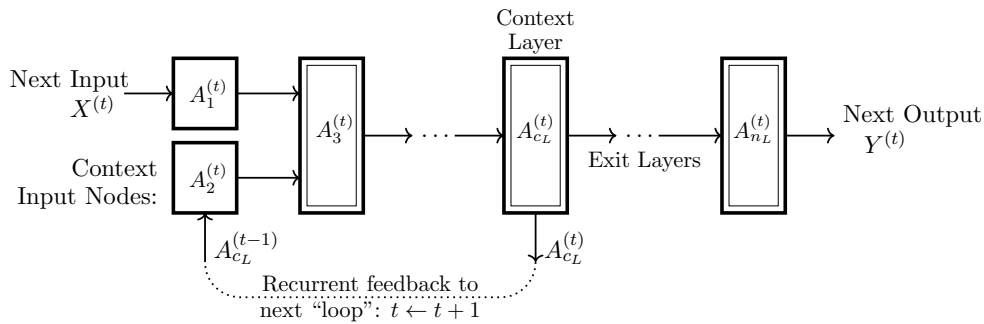
\begin{figure}[ht]
  \begin{center}
  \scalebox{.85}{
    \begin{tikzpicture}[xscale=1.1, yscale=1.1]
   \draw [black, ultra thick] (4.5,4.5) rectangle (5.4,5.5);
   \draw [black, ultra thick] (4.5,3.3) rectangle (5.4,4.3);
  \node at (5,5) {\begin{small}$A_1^{(t)}$\end{small}};
  \draw [->, thick] (3.8,5.0) -- (4.5,5.0);
  \node [above left] at (4,4.9) {Next Input};
  \node [below left] at (3.8,5.1) {$X^{(t)}$};
  \node at (5,3.8) {\begin{small}$A_2^{(t)}$\end{small}};
  \node [above left] at (4.4,3.7) {Context};
  \node [below left] at (4.4,3.8) {Input Nodes:};
  \draw [->, thick] (5.4,5.0) -- (6.3,5.0);
  \draw [->, thick] (5.4,3.8) -- (6.3,3.8);
  \draw [->, thick] (4.95,2.6) -- (4.95,3.3);
  \node [right] at (4.95,2.8) {$A_{c_L}^{(t-1)}$};
   \draw [black, ultra thick] (6.3,3.3) rectangle (7.2,5.5);
  \draw [black] (6.4,3.4) rectangle (7.1,5.4);
  \node at (6.8,4.5) {\begin{small}$A_3^{(t)}$\end{small}};
  \draw [->, thick] (7.2,4.4) -- (7.9,4.4);
  \node [right] at (7.9,4.4) {$\ldots$};
  \draw [->, thick] (8.5,4.4) -- (9.2,4.4);
   \draw [black, ultra thick] (9.2,3.3) rectangle (10.1,5.5);
  \draw [black] (9.3,3.4) rectangle (10.,5.4);
  \node at (9.65,4.5) {\begin{small}$A_{c_L}^{(t)}$\end{small}};
  \node [above] at (9.65,5.85) {\begin{small}Context\end{small}};
  \node [above] at (9.65,5.45) {\begin{small}Layer\end{small}};
  \draw [->, thick] (10.1,4.4) -- (10.8,4.4);
  \draw [->, thick] (9.65,3.3) -- (9.65,2.6);
  \node [right] at (9.65,2.8) {$A_{c_L}^{(t)}$};
  \node [right] at (10.8,4.4) {$\ldots$};
  \node [below] at (11.2,4.3) {\begin{small}Exit Layers\end{small}};
  \draw [->, thick] (11.4,4.4) -- (12.3,4.4);
  \draw [black, ultra thick] (12.3,3.3) rectangle (13.2,5.5);
  \draw [black] (12.4,3.4) rectangle (13.1,5.4);
  \node at (12.75,4.5) {\begin{small}$A_{n_L}^{(t)}$\end{small}};
  \draw [->, thick] (13.2,4.4) -- (13.9,4.4);
  \node [above right] at (13.9,4.4) {Next Output};
  \node [below right] at (14.2,4.6) {$Y^{(t)}$};
  
  \node at (7.3,2.3) {\begin{small}Recurrent feedback to\end{small}};
  \node at (7.3,1.9) {\begin{small}next ``loop'': $t\leftarrow t+1$\end{small}};
\draw [-,thick,dotted] (9.65,2.6) to [out=-90,in=0] (9.15,2.1)  to  (5.45,2.1) to [out=180,in=-90] (4.95,2.6);
    \end{tikzpicture}}  
\caption {Diagram showing dataflow in a Recurrent Neural Network (RNN).  Arrows show dataflow.  Each rectangle shows a layer of nodes in the neural network; the layers with only a single rectangle are those that make no transformation to the incoming data.  The data cycles around the network multiple times in ``loops'', each loop indexed by $\tloop$.  Algorithm \ref{alg:rnnDynamics} describes the process in greater detail.  The ``exit layers'' do any necessary post-processing on the data.  Extra shortcut connections, or repeated recurrent structure, may be present to obtain different RNN architectures.}
    \label{fig:rnnTopology}
  \end{center}  
\end{figure}

A simplified recurrent architecture is shown in Fig. \ref{fig:rnnTopology}.  This architecture consumes $\nloop$ input matrices $X^{(\tloop)}$, one at each time step $\tloop \in \{1,...,\nloop \}$, and produces $\nloop$ output matrices $Y^{(\tloop)}$.  
At each time step, data from an input matrix $X^{(\tloop)}$ enters the RNN from the left and propagates forwards in the usual manner.  When data reaches the ``context layer'', layer $\cl$, it loops back to the start of the RNN, and is combined with the next input matrix to go through the RNN again.   Data loops around the recurrent layers many times, each time also passing through the exit layers which perform some final post-processing on the data to  deliver the output matrices $Y^{(\tloop)}$.

Pseudocode is given in Alg. \ref{alg:rnnDynamics}.  In this notation, layer 0 is reserved for the bias nodes; layer 1 is for the input matrices $X^{(\tloop)}$, and layer 2 is for feedback received from the later context layer $\cl$.  Superscript numbers in brackets indicate the time step, $\tloop$.  

\begin{algorithm}[h!]
\caption{Recurrent NN Dynamics}
\label{alg:rnnDynamics}
\begin{algorithmic}[1]
\REQUIRE {On entry, require $\nloop$ input matrices ${X}^{(\tloop)} \in \Reals^{d_1\times\np}.$}
\STATE $A_\cl^{(0)}\leftarrow 0$ \COMMENT{Initial context units are zero}
\FOR{$\tloop=1$ to $\nloop$}
\STATE $A_0^{(\tloop)} \leftarrow [1\ 1\ \ldots \ 1]$ \COMMENT{Bias nodes}
\STATE $A_1^{(\tloop)} \leftarrow X^{(\tloop)}$ \COMMENT{$\tloop^{\mathrm{th}}$ input matrix.}
\STATE $A_2^{(\tloop)} \leftarrow A_\cl^{(\tloop-1)}$ \COMMENT{Feedback from context layer}
\FOR{$j=3$ to $\nl$}
\STATE $S_j^{(\tloop)} \leftarrow \layerWeightStack{j}\layerInputStack{j}^{(\tloop)}$ \label{line:rnnSums} 
\STATE $A_j^{(\tloop)} \leftarrow g(S_j^{(\tloop)})$ 
\ENDFOR
\STATE $Y^{(\tloop)} \leftarrow A_{\nl}^{(\tloop)}$  \label{line:rnnYassignment} \COMMENT{$\tloop^{\mathrm{th}}$ output matrix. $Y^{(\tloop)} \in \Reals^{d_\nl\times\np}$.}
\ENDFOR
\end{algorithmic} 
\end{algorithm}

Each input matrix $X^{(\tloop)}$ may itself contain a batch of several patterns (one in each column). Hence the matrices $A^{(\tloop)}_j$ and $S^{(\tloop)}_j$ have dimension $d_j \times \nb$.   

An appropriate loss function $\errorFunction$ would be chosen that is a function of some or all of the $Y^{(\tloop)}$ matrices, and then the gradient of this loss function with respect to the weights of the network, $\dEdw$, can be found by automatic differentiation, using for example, backpropagation through time \citep{backproptime90}, in execution time  $O(\nb\nloop\nw)$, where $\nw$ is the number of weights in the network.  Then, assuming weight-space is being used, an iterative optimizer would use this gradient information to tune $\vecW$, and train the network.

To incorporate target-space learning for a RNN, the intermediate objective is to make all the $S_j^{(\tloop)}$ coming from Alg. \ref{alg:rnnDynamics} match as closely as possible some given target matrices $T_j^{(\tloop)}$, for all time steps $\tloop$.  Hence, considering line \ref{line:rnnSums} of Alg. \ref{alg:rnnDynamics}, the objective is to choose a weight matrix $\layerWeightStack{j}$ so as to achieve, 
$$\layerWeightStack{j}\layerInputStack{j}^{(\tloop)}\approx T_j^{(\tloop)}\ \ \text{for all $1 \leq \tloop\leq \nloop$}\text{,}$$ 
or equivalently to achieve, as closely as possible,
$$\layerWeightStack{j}\layerInputStack{j}^{(:)}\approx T_j^{(:)},$$
where we have defined
\begin{subequations} \label{eqn:rnn:rolledUpTNotation}
\begin{equation}
\layerInputStack{j}^{(:)}\isDefinedToBe \begin{pmatrix}\layerInputStack{j}^{(1)}&\layerInputStack{j}^{(2)}&\ldots&\layerInputStack{j}^{(\nloop)}\end{pmatrix}
,\ 3 \leq j \leq \nl    
\label{eqn:rnn:rolledUpTNotationB} \end{equation} and     
\begin{equation}T_j^{(:)}\isDefinedToBe  \begin{pmatrix}T_j^{(1)}&T_j^{(2)}&\ldots&T_j^{(\nloop)}\end{pmatrix}
,\ 3 \leq j \leq \nl
\label{eqn:rnn:rolledUpTNotationT}
\end{equation}
\end{subequations}

The least squares solution to this is the same as in \eqref{eqn:WlayeredDiscreteFCCRealisation} and \eqref{eqn:BdaggerDefinition}: \begin{align}
\layerWeightStack{j} = T_j^{(:)}{\left(\layerInputStack{j}^{(:)}\right)^\pseudoinv}, \label{eqn:leastSquaresRecurrentEquation}
\end{align}however since this is a RNN, we now have the problem in that it is not possible to know the values of $\layerInputStack{j}^{(:)}$ until the network can by run by Alg. \ref{alg:rnnDynamics}; but that algorithm cannot be run until equation \eqref{eqn:leastSquaresRecurrentEquation} is solved.

To break out of this cyclic dependency, we can approximate using the ``optimistic'' cascade untangling (OCU), given by \eqref{eqn:mccApproximation}, and therefore just set:
\begin{align}
A_j^{(\tloop)} \leftarrow g\left(T_j^{(\tloop)}\right)\ \forall \tloop.\label{eqn:mccApproximationRNN}
\end{align}
This OCU step only needs doing on the context layer which feeds backward connections to the input layers.  For the rest of the layers, it is preferable to use the SCU method. Alg. \ref{alg:layeredRNNDiscreteRealisationFCC} shows how to do this in detail.  This algorithm calculates the weights of a RNN from a given list of target matrices $T_j^{(\tloop)}$, using the SCU method wherever possible, and the OCU method for the recurrent layer. The algorithm includes in line \ref{line:lrnnr:MCCcorrection} an attempt to correct the error introduced by the OCU step once the exit layers (shown in Fig. \ref{fig:rnnTopology}) are reached.

To modify the algorithm to a fully OCU method, then we would replace line \ref{line:lrnrr:Aequals} by equation \eqref{eqn:mccApproximationRNN}, and delete lines \ref{line:lrnrr:Sequals}  and \ref{line:lrnnr:MCCcorrection}.

\begin{algorithm}[h!]
\caption{Conversion of Targets to Weights for a RNN (using SCU)}
\label{alg:layeredRNNDiscreteRealisationFCC}
\begin{algorithmic}[1]
\REQUIRE {On entry, require $\nloopt$ input matrices $\overline{X}^{(\tloop)}  \in \Reals^{d_1\times \nbt}.$}
\STATE $A_0^{(\tloop)} \leftarrow [1\ 1\ \ldots \ 1]\ \forall \tloop$ \COMMENT{Bias nodes}
\STATE $A_1^{(\tloop)} \leftarrow \overline{X}^{(\tloop)}\ \forall \tloop$
\STATE $A_\cl^{(\tloop)} \leftarrow g\left(T_\cl^{(\tloop)}\right) \ \forall \tloop$ \label{line:lrnnr:AProjEquals}\COMMENT{Estimates $A_\cl^{(\tloop)}$ matrices by
OCU method.}
\STATE $A_2^{(:)} \leftarrow \begin{pmatrix} 0 & A_\cl^{(1)} &  A_\cl^{(2)} & \ldots &  A_\cl^{(\nloop-1)} \end{pmatrix}$ \COMMENT{Applies recurrent feedback from layer $\cl$ to layer 2.  Hence $A_2^{(:)}$ is a block shifted-right version of $A_\cl^{(:)}$.}
\FOR{$j=3$  to $\nl$}
\STATE $\layerWeightStack{j} \leftarrow T_j^{(:)}{\left(\layerInputStack{j}^{(:)}\right)^\pseudoinv}$ \label{line:lrnnr:Wequals} \COMMENT{Calculates weights to layer $j$}
\STATE $S_j^{(:)} \leftarrow \layerWeightStack{j}\layerInputStack{j}^{(:)}.$ \label{line:lrnrr:Sequals} 
\STATE $A_j^{(:)} \leftarrow g\left(S_j^{(:)}\right)$ \COMMENT{SCU method} \label{line:lrnrr:Aequals}
\IF {$j=\cl$}
\STATE Use the newly calculated $\layerWeightStack{j}$ matrices (for $3\leq j \leq \cl$) to run Alg. \ref{alg:rnnDynamics} (using $\overline{X}^{(\tloop)}$ as the input matrices), up to layer $\cl$, to obtain the true $\layerInputStack{\cl+1}^{(:)}$ matrices.  \COMMENT{This is an attempt to correct for the OCU estimation made in line \ref{line:lrnnr:AProjEquals}.} \label{line:lrnnr:MCCcorrection}
\ENDIF
\ENDFOR
\end{algorithmic} 
\end{algorithm} 

For the reasons discussed in Sec. \ref{sec:minibatching}, the content and length of the target-space input matrices, $\overline{X}^{(t)}$ for $t=1,\ldots,\nloopt$, may differ from the content and length of the weight-space input matrices ($X^{(t)}$ for $t=1,\ldots,\nloop$).

This algorithm merely outputs a set of weights of the RNN.  The RNN would then have to be run separately, using Alg. \ref{alg:rnnDynamics}, to obtain the set of output matrices $Y^{(\tloop)}$.

Since Alg. \ref{alg:layeredRNNDiscreteRealisationFCC}  defines the mapping from targets to weights, it is possible to calculate the learning gradient with respect to the targets (first going via $\dEdw$) using automatic differentiation, and hence train the RNN in target space. For example, if Alg. \ref{alg:layeredRNNDiscreteRealisationFCC} followed by Alg. \ref{alg:rnnDynamics} is passed to an auto-differentiation toolbox, then the toolbox will be able to correctly calculate $\dEdt$ by differentiation through both algorithms sequentially. Section \ref{sec:experiments} shows experiments which do this, with successful results.

The bottleneck in algorithmic complexity for Alg. \ref{alg:layeredRNNDiscreteRealisationFCC} is in forming the Gramian matrix $AA^T$, which will take ${\numinputs}^2\nbt \nloopt$ flops by direct multiplication.  This is similar to a full forward-unroll of the RNN with the input matrix $\overline{X}$.  Hence the relative complexity of running Alg. \ref{alg:layeredRNNDiscreteRealisationFCC} using $\overline{X}$, compared to Alg. \ref{alg:rnnDynamics} using input matrix $X$, is approximately $\nbt \nloopt/\nb \nloop$.  This motivates a choice of using a small value of $\nloopt$ where possible.  See Section \ref{sec:imdb_rnn} for an example.

\subsection{Application to Convolutional Neural Networks} \label{sec:cnnsTargetSpace}

Convolutional Neural Networks (CNNs) represent one of the most powerful modern deep-learning architectures and are particularly applicable to vision problems.  The key innovation of the convolutional neural network is the 2D-convolution operation: a smaller weight matrix is ``convolved'' (i.e. a sliding dot product is performed) with the source image to calculate the activations in the next layer.  The convolutional operation means the weight matrix connecting one layer to the next can be much smaller than that of a fully connected network; and also that this smaller group of weights, the convolutional ``kernel'', will be applied to multiple patches of the image.   This reuse helps in generalisation, and helps preserve spatial relationships in the image from one layer to the next.

A CNN network structure is usually comprised of a mixture of layer types - including one or more convolutional layers, one or more down-sampling (max-pooling) layers, flattening operations that reduce a tensor from rank 4 down to rank 2, and one or more regular fully-connected layers (as described in Section \ref{sec:algorithm}).  Further details of how these layers all work and are arranged with each other are given by \cite{lecun1998gradient}.  

In generating a target-space method for training a CNN, it is only the convolutional layers and fully-connected layers that have any weights, and so only those two layer types that need modifying.  

Each convolutional layer takes as input a 2D image, of size width$\times$height, with a third depth dimension representing a number of input channels.  Together with the batch size, $\nb$, this input image is a rank-4 tensor, of shape [$\nb$, input\_height, input\_width, input\_channels].  The convolutional kernel that acts on it is a rank-4 tensor of shape [kernel\_height, kernel\_width, input\_channels, output\_channels], and the layer's final output is a rank-4 tensor of shape [$\nb$, output\_height, output\_width, output\_channels].

The entire convolutional layer's operation can be split into 6 steps:
\begin{enumerate}
    \item Flatten the kernel to a 2-D matrix with shape [kernel\_height$\times$kernel\_width$\times$input\_channels, output\_channels].  Call this matrix $W$.
    \item Extract image patches from the input tensor, and reshape them, to form a patches matrix $A$ of shape [num\_patches, kernel\_height$\times$kernel\_width$\times$input\_channels], where num\_patches$=\nb\times$output\_height$\times$output\_width.  
    \item Multiply the kernel matrix $W$ by the patches matrix $A$, obtaining $S=WA$.\label{step:cnnMatrixMultiply}
    \item Add in the bias to $S$. \label{step:cnnBiasAdd}
    \item Reshape the result back into rank-4 tensor of shape [$\nb$, output\_height, output\_width, output\_channels] \label{step:cnnReshape4}
    \item Apply the activation function $g$. \label{step:cnnActivationFunction}
\end{enumerate}
    
To optimise this process, so as to be able to easily modify it for target-space training, we first combine the bias addition of step \ref{step:cnnBiasAdd} with the matrix multiplication of step \ref{step:cnnMatrixMultiply}.  This can be achieved by adding an extra row of 1s into $A$, as was done in equation \eqref{eqn:AmatrixWithBiasCombined}, and an extra column of weights to $W$, as was done in equation \eqref{eqn:WmatrixWithBiasCombined}.

Then we need a target matrix $T$ of the same dimension as $S$ in line \ref{step:cnnMatrixMultiply}. Given this target matrix and the matrix $A$, we can derive the weights which best achieve the targets using the same least-squares process as with equation \eqref{eqn:WlayeredDiscreteFCCRealisation}, i.e. $W=TA^\pseudoinv$.

This derived weight matrix $W$ is then used to calculate the actual product $S=WA$, and steps \ref{step:cnnReshape4} and \ref{step:cnnActivationFunction} (the reshape and activation function) are applied, completing the convolutional layer's behaviour.

The fully-connected layers are handled with their own target matrices and least-squares solution, as in Alg. \ref{alg:mappingTargetsToWeightsSCU}. The rest of the layer-types in the CNN are unchanged - down-sampling does not use any targets (or weights), and nor does the reshape operation.

Automatic differentiation can be used to compute the necessary learning gradients.  

The algorithmic complexity for the target-space CNN layer is derived in Appendix \ref{sec:appendixCnnComputationalComplexity}, and is shown to be slower than the corresponding weight-space CNN layer by a factor which is bounded above by approximately $(3(\kernelheight \kernelwidth)+1)\nbt/\nb$, where $\kernelheight$ and $\kernelwidth$ are the kernel height and width, respectively.  This is not a constant bound, even when $\nbt=\nb$, unlike that found for the fully-connected network.\footnote{\label{footnote:optimisation}In future work, it is possible to remove this numerator factor of $\kernelheight \kernelwidth$, since with a stride-length of 1 there is significant overlap between patches in the matrix $A$, and therefore optimisations can be made when forming $AA^T$.}     Hence there is an incentive in target space to choose CNN architectures with smaller kernel matrices, or to only use a subset of patches when forming the pseudoinverse matrix.  In the CNN architectures used in the experiments of Section \ref{sec:cnnExperiments}, the ratio is empirically found to be around 7 (with a 3-by-3 kernel), which is considerably better than the theoretical upper-bound.  Part of this improvement might be down to the fact that the backward pass of automatic differentiation can reuse the expensive matrix products and inverses computed in the forward pass.

This completes the description of how to use target space with a conventional CNN architecture.

\section{Experiments} \label{sec:experiments}

In this section we show the performance of the target-space method on the Two-Spirals benchmark problem, and on four classic small-image vision benchmark problems for convolutional neural networks, and then we demonstrate the target-space method on some bit-stream manipulation tasks and a sentiment-analysis task for recurrent neural networks.

The experiments show the effectiveness of the target-space method, in ability to train deep networks and produce improved generalisation. There are improved generalisation results on the CNN vision benchmarks compared to the equivalent weight-space method applied to the same CNN architecture.  In the recurrent network tasks, it shows the target-space method being able to solve problems with long time-sequences, which appear to be intractable in weight space.

All experiments were implemented using Python and Tensorflow v1.14 on a Tesla K80 GPU.\footnote{Source code for experiments is available at \url{https://github.com/mikefairbank/dlts_paper_code}}  Shading in graphs indicates 95\% confidence intervals as calculated by the Python Seaborn package.   

\subsection{Two-Spirals Experiments} \label{sec:twoSpiralsExperiments}
\begin{figure}[h]
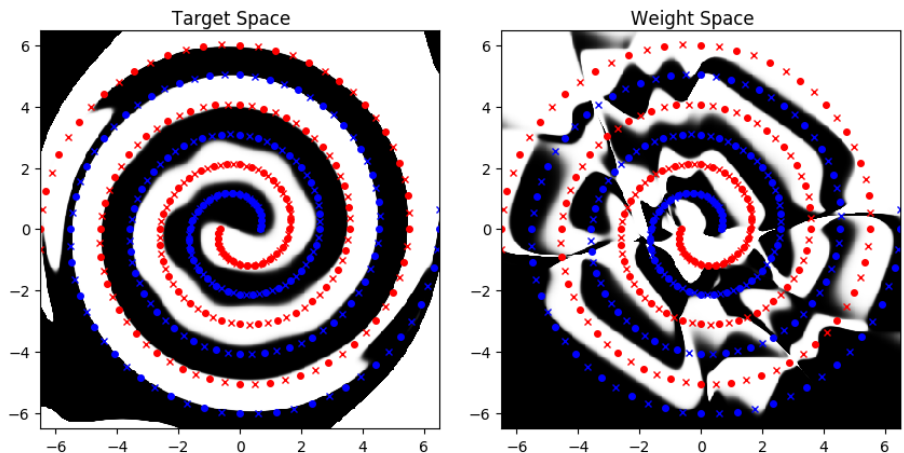

\centering
\includegraphics[width=6cm]{\detokenize{trained_net_t.png}} 
\includegraphics[width=6cm]{\detokenize{trained_net_w.png}} 
\caption{Typical results for the two-spirals trained network, after 4,000 Adam iterations; target space versus weight space.    Red/blue crosses denote test set; circles denote the training set.  Grey-scale background indicates network output for the given $(x,y)$-coordinate input. Smoothness of the target-space result shows how successful generalisation is more likely.} \label{fig:resultsTwoSpiralsOutputImage} 
\end{figure}

The Two-Spirals classification problem consists of 194 two-dimensional training points, arranged in two interleaving spiral shapes, corresponding to the two output classes, each spiral revolving through three complete revolutions.   The training and test sets are shown in Fig. \ref{fig:resultsTwoSpiralsOutputImage}.  The test set was created as the angular midpoints between consecutive training points.

A layered network architecture was used, with dimensions 2-5-5-5-2, and with all shortcut connections, following \cite{riedmiller93direct}.  The cross-entropy loss function was used for training, and the $\tanh$ activation function used on all hidden layers, with softmax on the output layer.

Fig. \ref{fig:resultsTwoSpiralsOutputImage} shows the output function of two trained networks, mapped to a single scalar output, and visually indicates that the solutions attained in target space are smoother and capture the essence of the problem better than in weight space.\footnote{Although it should be noted that Levenberg Marquardt and conjugate gradient training can produce similarly nice solutions as the left figure.} 

\begin{figure}[h]
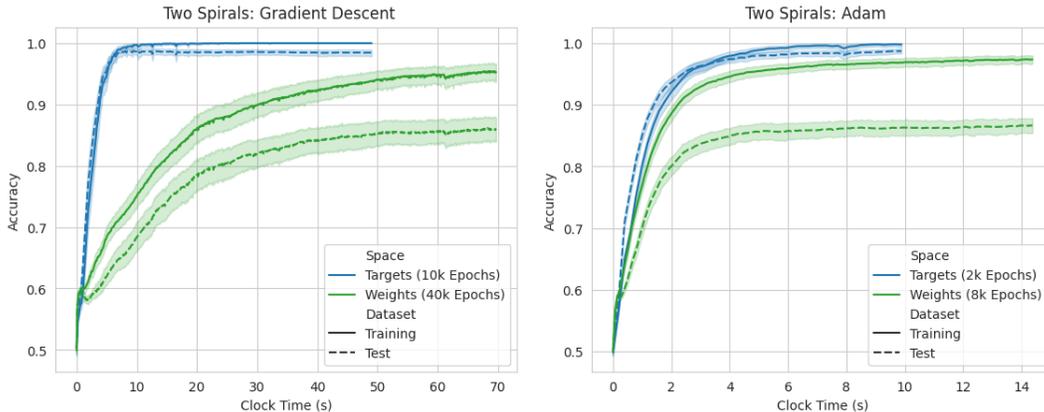

\centering
\includegraphics[width=7cm]{\detokenize{two_spirals_gradient_descent_experiments_summary_clock_time.png}} 
\includegraphics[width=7cm]{\detokenize{two_spirals_adam_experiments_summary0.01_clock_time.png}} 
\caption{Results for Two-Spirals learning, using Batch Gradient Descent (on left) and Adam optimiser (on right).}\label{fig:resultsTwoSpiralsGD} 
\end{figure}

Fig. \ref{fig:resultsTwoSpiralsGD}-left shows the problem being solved using gradient-descent with optimal learning rates empirically determined as $\eta=10$ for target space and $\eta=0.1$ for weight space.  The results show that with optimal learning rates, the target-space algorithm can fully learn the two-spirals problem's training set, and generalise well to the test set, in around 1,000 epochs; compared to around 40,000 epochs for weight space to mostly learn the training set only.  It does not seem possible to generalise as well to the test set in weight space, likely due to the unevenness appearing in Fig. \ref{fig:resultsTwoSpiralsOutputImage}-right. 

Fig. \ref{fig:resultsTwoSpiralsGD}-right shows results when the Adam optimiser was used, and shows a similar outcome.  The learning rate used was 0.01, which was found to be beneficial to both target space and weight space on this problem. In this problem, the target-space gradient descent converges to a solution in fewer epochs than Adam in weight space.

These results all seem consistent with the target-space motivation for making the loss-function surface smoother, and the minima commonly found lead to better generalisation.

In our implementation the processing time was on average 3.5 times longer for each target-space training iteration compared to each weight-space iteration.  In all experiments, the full data-set was used in all training batches ($\nb=\nbt=194$).  With target space, $\lambda=0.001$ was used for equation \eqref{eqn:BdaggerDefinition}, and initial targets were randomised using a truncated normal distribution with $\sigma=1$, followed by the projection given by \eqref{eqn:initialTargetsProjection}.  For weight-space learning, the weights were randomised using the method of \cite{glorot2010understanding}.

\begin{figure}[h]
\centering
\includegraphics[width=9cm]{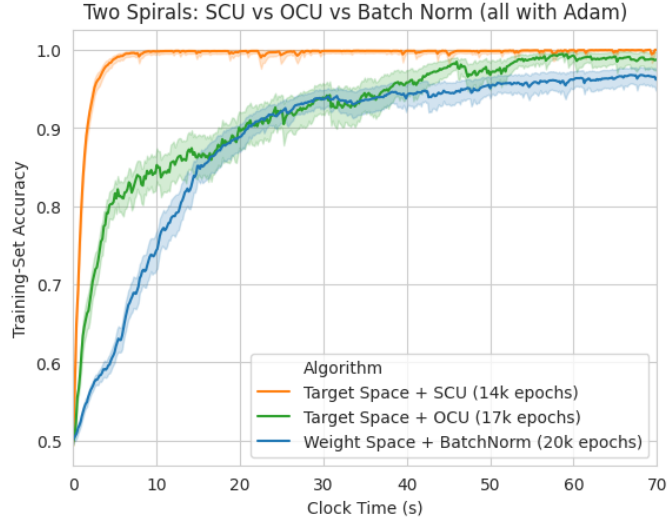}
\caption{Results for Two-Spirals learning, using Adam Optimiser, comparing two forms of target space: Optimistic Cascade untangling (OCU)  versus Sequential Cascade untangling (SCU), and against Batch Normalisation in weight space.}\label{fig:resultsTwoSpiralsFCC_MCC_bn} 
\end{figure}

Fig. \ref{fig:resultsTwoSpiralsFCC_MCC_bn} shows the effectiveness of the Sequential Cascade untangling (SCU) variant against the Optimistic Cascade untangling (OCU) target-space algorithm (described in Section \ref{sec:layeredRealisationMain}), and indicates that the SCU method is more stable and effective than the OCU method.  

The same graph also shows that Batch Normalisation does not seem to help on this problem and network size, and in fact performs worse in weight space than without batch normalisation.  Batch normalisation does significantly help though in the CNN experiments described in the next subsection.

\begin{figure}[h]
\centering
\includegraphics[width=7.5cm]{\detokenize{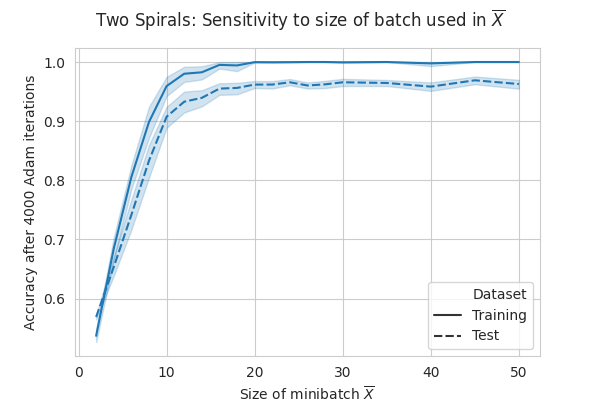}} 
\includegraphics[width=7.5cm]{\detokenize{two_spirals_sensitivity_lambda_4000.png}}
\caption{Sensitivity of the Target Space algorithm to algorithm hyper-parameters $\nbt$ and $\lambda$. }\label{fig:hyperParameterSensitivity1} 
\end{figure}

Fig. \ref{fig:hyperParameterSensitivity1} demonstrates the sensitivity of the target-space algorithm to two of its key hyper-parameters.  The left diagram shows that reducing $\nbt$, the number of patterns appearing in $\overline{X}$ (see Section \ref{sec:minibatching}), reduces the representation capability of the weights generated by equation \eqref{eqn:WlayeredDiscreteFCCRealisation}.  In each experimental trial, a random subset of $\nbt$ columns of $X$ was chosen to form $\overline{X}$.  The results show that as $\nbt$ reduces below the size of the narrowest network layer (which is 17 in this network), the weight matrices generated from the targets become low-rank, and it is no longer possible to fully learn the training set. 
Fig. \ref{fig:hyperParameterSensitivity1}-right shows that as the $\lambda$ used in \eqref{eqn:BdaggerDefinition} increases, the ability of the algorithm also reduces (see Section \ref{sec:choiceoflambda}).  Furthermore, with $\lambda \ll 10^{-4}$ the algorithm stopped due to numerical errors causing non-invertible matrices to appear.
\subsection{CNN Experiments}\label{sec:cnnExperiments}

In this set of experiments we train convolutional neural networks on the following four classic small-image classification problems:
\begin{itemize}
    \item The MNIST digit dataset: 60,000 training samples of 28-by-28 grey-scale pixellated hand-written numeric digits, each labelled from 0-9, and a test set of 10,000 samples \citep{lecun2010mnist}.
    \item MNIST-Fashion dataset:  60,000 28x28 grayscale images of 10 labelled fashion categories, along with a test set of 10,000 images \citep{xiao2017fashion}.
    \item CIFAR10 dataset: 50,000 32x32 colour training images, labelled over 10 categories, and 10,000 test images \citep{krizhevsky2009learning}.
    \item CIFAR100 dataset: 50,000 32x32 colour training images, labelled over 100 categories, and 10,000 test images  \citep{krizhevsky2009learning}.
\end{itemize}

All of these datasets were used as training data without any modification to the training images. For example, we did not use any data-augmentation techniques, such as image rescaling and distortion, which are known to help improve neural-network performance (and to be necessary to achieve state-of-the art classification performance).

The networks used here all had six compound convolutional/pooling layers, each of which consisted of a convolutional operation (with a square kernel of size $m\times m$, applied with stride-length 1 with ``same'' padding, and $c$ output channels) followed by an application of the activation function, followed by (possibly) an application of max-pooling (with a square kernel of size $k \times k$, and applied with stride-length $k$). Each max pooling operation of side length $k$ reduces the side-length of the image by factor $k$.  Hence each compound convolutional layer can be summarised by a 3-tuple $(m,c,k)$, with $k=1$ if no max-pooling is used.     Using this 3-tuple notation, the network architectures considered are listed in Table
\ref{table:cnn_architectures}.  

After the convolutional layers, the layer output is flattened, and then passed through a number of fully-connected (dense) layers, as described in Table
\ref{table:cnn_architectures}.

\begin{table*}[ht]
\centering
\begin{tabular}{|r||p{9cm}|l|l |}
\hline
Benchmark   &  Convolutional Layers  &\\ 
Problem &(Convolution size - Number of channels - Max Pool size)  &Dense Layers \\\hline \hline 
MNIST&(3-16-1)-(3-16-2)-(3-32-1)-(3-32-2)-(3-64-1)-(3-64-2)&128-10\\
MNIST-Fashion&(3-16-1)-(3-16-2)-(3-32-1)-(3-32-2)-(3-64-1)-(3-64-2)&128-10\\
CIFAR-10&(3-32-1)-(3-32-2)-(3-64-1)-(3-64-2)-(3-128-1)-(3-128-2)&128-10\\
CIFAR-100&(3-32-1)-(3-32-2)-(3-64-1)-(3-64-2)-(3-128-1)-(3-128-2)&512-128-100\\
\hline
\end{tabular}
\caption{Convolutional Network Architectures considered for MNIST Problem}
\label{table:cnn_architectures}
\end{table*}

All non-final layers used the ``leaky-relu'' activation function \citep{maas2013rectifier} defined by, 
\begin{equation}
    \lrelu(x)=\max(x,0.2x), \label{eqn:lrelu}
\end{equation}
and the final layer used softmax activation.  Leaky-relu was found to slightly be better than the $\relu$ function, since it leaves fewer zeros in the activations which can potentially stall learning after the weights are initially randomised; and also can potentially make the Gramian matrix in \eqref{eqn:BdaggerDefinition} low rank.

The networks were trained with the cross-entropy loss function and the Adam optimizer, with learning rate $0.001$ for weight-space learning, and $0.01$ for target-space learning.  Mini-batches of size $\nb=100$ were randomly generated at each iteration, for computing the $\dEdw$ gradient. A fixed mini-batch of size $\nbt=100$ was used for the targets' input matrix $\overline{X}$.

In weight space, the weight initialisation used magnitudes defined by \cite{he2015delving}, which are derived to work well with $\lrelu$.  In target space, the targets values were all initially randomised with a truncated normal distribution with standard deviation 0.1, followed by the projection operation given by \eqref{eqn:initialTargetsProjection}. $\lambda=0.1$ was used in equation \eqref{eqn:BdaggerDefinition}.

Results are shown in Table \ref{table:cnnDatasetResults}. The results show the target-space method helping generalisation performance, both with and without dropout \citep{srivastava2014dropout}, and when comparing against weight space both with and without batch-normalisation; and with ensemble architectures.  The benefit of target space is noticeable in the latter 3 benchmark problems; mostly so in the most challenging benchmark problem, i.e. CIFAR100.

The two CIFAR problems were given a time budget of 24 GPU hours to train each network.  This allowed approximately 640 epochs in target space, and 5300 epochs in weight space (lowering to 4000 epochs when batch-norm was used).  The two MNIST problems received a 8 GPU-hour time budget, resulting in approximately 480/3000/2400 epochs for target-space/weight space/BN, respectively. Hence roughly seven times more processing time was required per epoch for the target-space algorithms compared to the weight-space algorithms.

\begin{table*}[ht]
\begin{small}
\hspace{-1cm}

\vspace{1cm}

\hspace{-1cm}
\begin{tabular}{|l ||c |c| c| c |}
\hline
Algorithm (no dropout) &  MNIST & MNIST-Fashion & CIFAR-10 & CIFAR-100 \\ \hline 
Weight Space 				&$99.26 (\pm0.01)\%$	& $91.6 (\pm0.1)\%$	& $77.9 (\pm0.2)\%$	& $40.3 (\pm0.7)\%$\\
Weight Space + Batch Normalisation	 &$\textbf{99.41} (\pm0.04)\%$	& $91.6 (\pm0.2)\%$	& $80.7 (\pm0.2)\%$	& $46.5 (\pm0.6)\%$\\
Target Space				&$99.29 (\pm0.04)\%$	& $\textbf{92.2} (\pm0.2)\%$	& $\textbf{82.6} (\pm0.3)\%$	& $\textbf{50.5} (\pm0.2)\%$\\
\hline
Algorithm (with dropout) &  MNIST & MNIST-Fashion & CIFAR-10 & CIFAR-100 \\ \hline 
Weight Space 				&$99.50 (\pm0.06)\%$	& $93.12 (\pm0.01)\%$	& $82.90 (\pm0.09)\%$	& $52.22 (\pm0.04)\%$\\
Weight Space + Batch Normalisation	 &$\textbf{99.58} (\pm0.01)\%$	& $\textbf{94.07} (\pm0.09)\%$	& $86.70 (\pm0.01)\%$	& $59.7 (\pm0.2)\%$\\
Target Space				&$99.55 (\pm0.03)\%$	& $93.7 (\pm0.1)\%$	& $\textbf{87.4} (\pm0.1)\%$	& $\textbf{60.4} (\pm0.1)\%$\\

\hline
Algorithm (with dropout + ensemble) &  MNIST & MNIST-Fashion & CIFAR-10 & CIFAR-100 \\ \hline 
Weight Space 				&99.49 \%	& 93.99 \%	& 85.43 \%	& 56.85 \%\\
Weight Space + Batch Normalisation	 &99.6 \%	& \textbf{94.5} \%	& 88.19 \%	& 62.51 \%\\
Target Space				&\textbf{99.62} \%	& 94.34 \%	& \textbf{88.81} \%	& \textbf{63.24} \%\\

\hline
\end{tabular}

\end{small}
\caption{Test-Set Accuracies for CNN Experiments, on Standard Datasets}
\label{table:cnnDatasetResults}
\end{table*}

When dropout was used, it was applied with a dropout probability of 0.2 to all non-final dense layers, and all even-numbered convolutional layers.  The results show that dropout provides useful benefit to both weight-space learning and target-space learning. 

When dropout was used in target space, dropout was independently applied during both the feed-forward algorithm used to calculate $\dEdw$ using the mini-batch input matrix $X$, and the feed-forward algorithm to map from target space to weight space using the fixed input matrix $\overline{X}$.\footnote{Generalisation results were noticeably worse if dropout in target space was applied to either one of these two stages without the other.}

When batch normalisation was used, it was applied to every convolutional layer and to every non-final dense layer.  Batch normalisation is only applicable to weight-space learning.  In target space learning, the targets for each layer already define the batch mean and standard-deviation which batch normalisation hopes to specify; making the combination of batch normalisation with target space redundant.

The error margins in Table \ref{table:cnnDatasetResults} are calculated as the standard-deviations of just two trials; but are sufficiently small to convey the trend adequately.  

When the ensemble of networks were used, the outputs of the two networks created in the two trials were averaged after softmax.  Ensemble networks can usually generalise better than any of their constituent networks individually, assuming the outputs of the constituent networks are somewhat independent of each other.  In this scenario the independence comes from different initial randomisation, different shuffling of mini-batches, and different choices of the $\overline{X}$ matrix used by the target-space algorithm.  The results show that target space and weight space are assisted by using such an ensemble; even one comprised of only two networks.

\subsection{Bit-Stream Recurrent Neural-Network Experiments} \label{sec:rnnExperiments}

In this section we describe two recurrent neural-network experiments regarding remembering and manipulating streams of bits.

The first experiment is to memorise and recall a random stream of bits.  The RNN receives a new random bit at every time step $t$, and must output the bit it saw at the time step $t-N$, where $N$ is the delay length.  As the delay length is increased, the problem gets harder, since more bits must be memorised.  

For example if the delay length is $N=2$, and the RNN receives a bit stream such as ``1,1,1,1,0,1'' (with most recent bits appearing at the right) then the RNN is expected to produce an output stream ``-,-,1,1,1,1''.  (The first two outputs in the sequence, each indicated by here ``-'', are ignored, since the delay length in this example is 2.)

The neural network has architecture $1-(N+3)-2$, with the hidden layer being fully connected to itself with recurrent connections (corresponding to setting $\cl=3$ in Fig. \ref{fig:rnnTopology}).   The hidden layer used $\tanh$ activation functions, and the final layer used softmax with cross-entropy loss function.  The loss function was made to ignore the first $N$ outputs in the stream (since these are undefined).

The $N+3$ recurrent hidden nodes are enough to allow the network to remember the most recent $N$ bits (with 3 spare nodes to add a little flexibility in solution), as required; for example the RNN could learn manipulate the remembered bits with a rotate-right bit-wise operation, so as to successfully queue and recall the bits, and forget about bits older than $N$. 

A batch size of 8,000 random bit streams of length $\nloop=N+50$ was used to train the network. Random mini-batches of size $\nb=100$ were used during each training iteration.  A fixed mini-batch of size $\nbt=100$ with $\nloopt=\nloop$ was used for the target-space matrices $\overline{X}^{(t)}$.  

In weight space, the weight initialisation used magnitudes defined by \cite{glorot2010understanding}.  In target space, the targets values were randomised with a truncated normal distribution with standard deviation 1, followed by a projection by equation \eqref{eqn:initialTargetsProjection}.  This projection step seemed to improve results for the target-space experiments.

The networks were trained with 50,000 iterations of Adam optimiser, with learning rate 0.001 for both weight-space and target space, and with $\lambda=0.1$ for target space.  

A result was considered a success if a classification accuracy $\geq 99\%$ was achieved on the test set at any training iteration; otherwise it was a failure.

\begin{figure}[h]
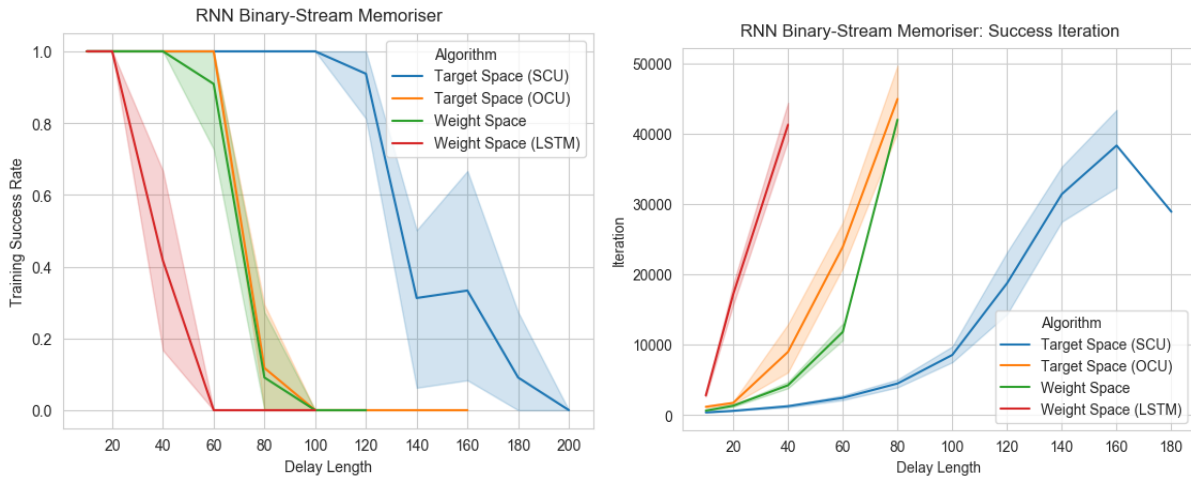

\centering
\hspace{-0.5cm}
\includegraphics[width=8cm]{\detokenize{delay_bit_problem_success_rate.png}}
\includegraphics[width=8cm]{\detokenize{delay_bit_problem_iteration.png}}
\caption{Memorisation of a delayed binary stream of bits using a RNN.  The left graph shows the ratio of trials which were successful in correctly learning $>99\%$ of the output bits correctly (in a test set).  The right-hand graph shows, for those successful trials, the average iteration number at which success was first achieved.}\label{fig:resultsRNNBitMemorisation}  
\end{figure}

Results are shown in Fig.  \ref{fig:resultsRNNBitMemorisation} for various delay lengths.  They show that the  target-space method is able to learn sequences with a delay length of around two to three times as long as the weight-space methods are capable of, with a significantly less steep rise in the number of training iterations required for success; and that the target-space SCU method is significantly stronger than the target-space OCU method.

For comparison, an extra experiment was made using an LSTM network.  Here the $N+3$ hidden nodes were replaced by $N+3$ LSTM memory cells.  The LSTM network was trained in weight space, again using Adam for 50,000 iterations.  Results are shown in the same Fig. \ref{fig:resultsRNNBitMemorisation}.  This trial shows that the LSTM network does not seem to help in solving this problem in weight-space.   

In a second RNN experiment, we modify the task from pure mermorization into one of binary addition.  In this experiment, the target output is the binary sum of the stream of bits with the $N$-step delayed stream.  To ease binary addition, the stream is assumed to arrive in bit-wise little-endian form.

For example, if $N=2$, and the bit stream received is ``1,0,1,1,0,1'', then the target output stream that the RNN must learn is ``-,-,0,0,0,1'', which is calculated by binary addition: 1101+1011=00011.  Here the target output stream terminated before the final carry bit could be delivered, so only the 0001 remained.

As this problem was slightly harder than the previous one, since the relationship between the target-bit sequence and the past sequence is quite well disguised (the relationship has similarities to a delayed XOR problem but there is also a hidden carry-bit process to discover), we gave the recurrent network $N+5$ hidden recurrent nodes, i.e. two more than previously.  

\begin{figure}[h]
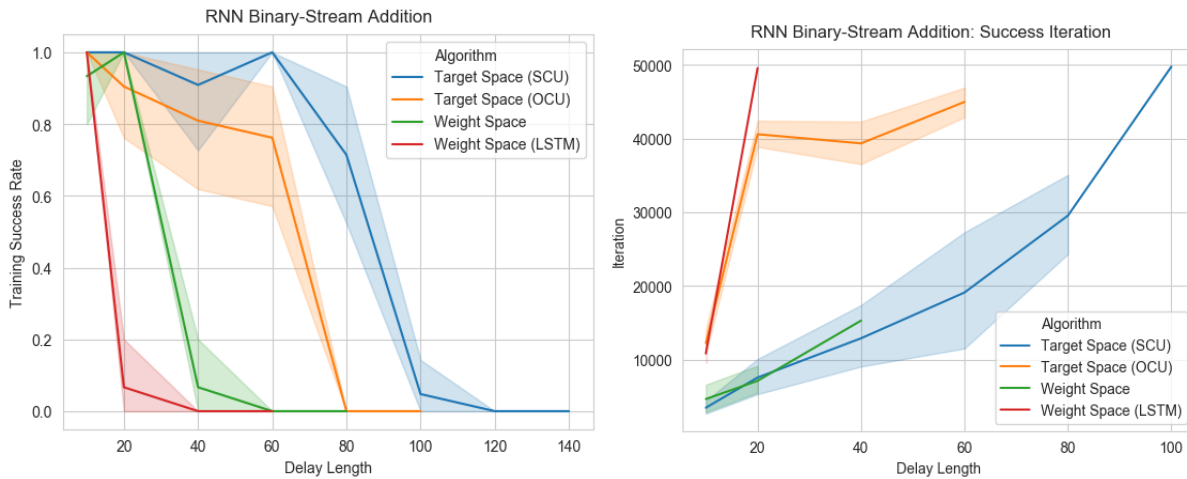

\centering
\hspace{-0.5cm}
\includegraphics[width=8cm]{\detokenize{delay_bit_problem_success_rate_adder.png}}
\includegraphics[width=8cm]{\detokenize{delay_bit_problem_iteration_adder.png}}
\caption{
Addition of a delayed binary stream of bits using a RNN.}\label{fig:resultsRNNBitAdder}
\end{figure}

Results are shown in Fig.  \ref{fig:resultsRNNBitAdder}.  The experimental conditions are otherwise unchanged from the previous RNN experiment.

In this experiment the strength of the target-space methods are again shown, with the SCU method again being capable of coping with delay lengths two to three times as long as the weight-space methods, and with better scaling of the number of iterations required. The strength of the SCU method's results confirms the value of lines \ref{line:lrnrr:Aequals} and \ref{line:lrnnr:MCCcorrection} in Alg. \ref{alg:layeredRNNDiscreteRealisationFCC}, when compared to the OCU method.

In both of these RNN experiments, the SCU method significantly beats the LSTM network.  It therefore seems that the exploding-gradients problem (which target-space networks are designed to address) is more significant in this problem than the vanishing-gradients problem (which LSTM networks are designed to address).  A complication in making this comparison is that Adam was used.  Adam might have been picking up and aggressively accelerating tiny components of the gradients in target space, thus counteracting the vanishing gradients and helping the target-space methods compete with LSTM.  Possibly in a more noisy problem environment, it will not be possible to accelerate such tiny gradients due to the low signal-to-noise ratio.  In that case a combination of LSTM plus target space could be attempted.

\subsection{RNN Movie-Review Sentiment Analysis} \label{sec:imdb_rnn}
In this final experiment we trained a RNN to solve the natural-language processing task of sentiment analysis for 50,000 movies reviews from the Internet Movie Database (IMDB) website.  In this binary classification task, each review is labelled as either positive or negative.  The dataset was obtained from the Tensorflow/Keras packages, with a 50-50 training/test-set split, using options of only including the top 5000 most frequent words, and padding/truncating all reviews to a length of 500 words each.

A word-embedding vector of length 32 was used to encode each word from the vocabulary of size 5000 \citep{bengio2003neural,mikolov2013linguistic}.  Once each word is converted into an embedded vector, the neural-network architecture is the same as in the previous experiment, but with 32 inputs, 100 nodes in the recurrent layer, and two output nodes.  Each embedded word of a review is fed to the RNN one-by-one, making the sequence length $\nloop=500$.  Only the final output matrix of the neural network, $Y^{(500)}$, is observed.

\begin{figure}[h]
\centering
\includegraphics[width=8cm]{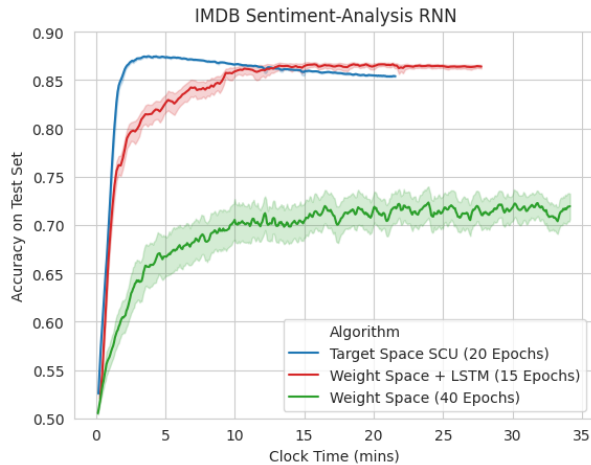}
\caption{Results for Movie Sentiment Analysis RNN Problem.}\label{fig:imdb_results}
\end{figure}

Results are shown in Fig. \ref{fig:imdb_results} and are summarised in Table \ref{table:imdb_results}, and show that the target-space method's performance slightly exceeds that of the LSTM network, and significantly exceeds ordinary neural networks trained in weight space.

\begin{table*}[ht]
\centering
\begin{tabular}{|r||r|r|}
\hline
Algorithm /   &  Best Test & Average GPU time  \\ 
Network Type  &  Accuracy & per Epoch (s)   \\ \hline \hline
Weight Space &	 $79.0 (\pm5.2)\%$ & 51.3\\
Weight Space + LSTM &	 $87.3 (\pm0.6)\%$ & 111.3  \\
Target Space&	 $87.7 (\pm0.2)\%$ & 64.9 \\
\hline
\end{tabular}
\caption{Results for Movie Sentiment Analysis RNN Problem}
\label{table:imdb_results}
\end{table*}

All neural networks were trained using Adam with learning rate 0.001, and mini-batch sizes of $\nb=40$. The target-space algorithm used $\lambda=0.001$. Weights and targets were initially randomised as in the previous subsection.  Word embeddings were also initially randomised (using a normal distribution with $\mu=0$ and $\sigma=0.1$).  Hence all weight and target matrices, and the embedding vectors, were learned in an end-to-end training process.

To customise the target-space method to handle word embeddings efficiently, a fixed sequence of target-space input matrices $\overline{X}^{(t)}$ was chosen, for a sequence length of just $\nloopt=60$, and mini-batch size $\nbt=40$. For efficiency, it was chosen that these matrices would represent some already-embedded word sequences.  Hence each matrix $\overline{X}^{(t)}\in \Reals^{ 32 \times 40}$, for $t=1,\ldots,60$.  Each of the $\overline{X}^{(t)}$ matrices was generated using a uniform random distribution in the range [-1,1], and then held constant throughout training.  The lower sequence length $\nloopt=60$ improves the algorithmic complexity factor (given at the end of Section \ref{sec:rnnsTargetSpace}), and results in a more competitive target-space training time in Table \ref{table:imdb_results}.
Even though this sequence length ($\nloopt=60$) was less than the true sequence length ($\nloop=500$), the combination of fixed matrices $\overline{X}^{(t)}$ and target matrices $T_j^{(t)}$ provide enough information to define the weight matrices $\layerWeightStack{j}$ unambiguously using Alg. \ref{alg:layeredRNNDiscreteRealisationFCC}; even though $\overline{X}^{(t)}$ are fixed random matrices, and therefore do not conform to any valid  movie-review style of writing. Hence the learning gradient $\dEdw$ (which now includes the gradient of the learnable embedding matrix, $W_{embed}\in \Reals^{5000\times 32}$), can be calculated in weight space, using the full sequence lengths (500), and then converted to a target-space gradient $\dEdt$, {using Alg. \ref{alg:layeredRNNDiscreteRealisationFCC}, followed by automatic differentiation.  To compute the gradient $\fracpartial{\errorFunction}{W_{embed}}$ in target space, in order to optimise those learnable variables too, we just used its value in weight space, without any modification.

\section{Conclusions}
\label{sec:conclusions}

The target-space method provides an alternative search space in which to train deep and recurrent neural networks.  The theory and experiments indicate that the loss-function surfaces being optimised are indeed smoother and 
easier to optimise in target-space than in weight space.  This increased smoothness potentially leads to easier solution of problems and potentially leads to better generalisation capabilities in the final neural networks produced.

Using target space comes at an added computational expense.  In fully connected networks, where the batch sizes for $X$ and $\overline{X}$ roughly match, this is usually a modest constant cost of approximately 3 or 4 times as much computation per training iteration.  With CNNs it can be more, being around 7 times in the CNN architectures considered in this paper, and more so if wider convolutional kernels are used.  With the RNN experiments, which can be considered as extremely deep and narrow networks, the timings were of similar order of magnitude between weight space and target space. It is hoped that by careful choice of architecture, focusing on deeper networks with narrower hidden layers \citep[possibly with several narrower layers running in parallel, which has already been proven as a powerful design by][in their ``ResNeXt'' CNN design]{xie2017aggregated}, and avoiding pattern-by-pattern learning, these costs can be minimised.

It has been shown how to combine mini-batching with target space.  The lack of mini-batching has historically been a major Achilles heel in the adoption of some previous sophisticated optimisers (for example conjugate gradients or Levenberg-Marquardt), with very large datasets.

Target-space methods are particularly promising in recurrent neural-network environments.  In the examples given, problems with sequence lengths that were previously intractable have been solved, and the LSTM results were surpassed in a natural-language problem. This is despite the fact that LSTM networks have extra features, such as memory gates, which make the learning task easier, yet the target-space learning has still managed to make ordinary RNNs outperform them.    In the feed-forward problems given, target space has consistently produced better generalisation in deeper neural networks.

The theoretical motivation for target space, in that using targets should be able to untangle the cascades of changes caused during training, with a beneficial outcome, appears to be feasible.  Hence target space aims to directly address the recognised ``exploding-gradients'' problem which exists in deep learning.

Regarding a hypothetical future of neural networks being able to produce simple programs similar to those formed by human programmers, we hypothesise that whenever a neural network gets to a really interesting point of training, then the neural activations will often all be very close to their firing thresholds, and the exploding-gradients problem becomes really significant in blocking further learning.  For example if the neural network training process had somehow successfully managed to build a series of interlocking XOR gates, which were almost all working well together so as to implement a conventional computer program out of those logic gates, then the scrambling of behaviour from any potential infinitesimal weight change will always make learning destabilise in weight space.  The target-space approach is designed to be helpful in these circumstances, and would seem to have more chance of making further progress than a simple weight-space search would.

Our experimental results with recurrent neural networks over long time sequences combined with data-processing outperform the equivalent LSTM networks.  Hence it seems that in those problems at least, the exploding-gradients problem is more significant than vanishing gradients; at least when Adam is allowed to accelerate the small gradients in target space.  This is particularly paradoxical when it is noted that the objective of the target-space cascade untangling is to dampen down learning gradients even more, thus amplifying the vanishing-gradients problem.    

Many significant deep-learning innovations exist in prior published work.  These include the closely-related method of batch normalisation, plus modern activation functions, optimisers, and weight-initialisation techniques.  Many of these are more computationally efficient than target space, but are maybe slightly less effective; and some can be combined with target space.    Sophisticated neural architectures, such as LSTM, CNNs, and more recently, attention models, Differentiable Neural Computers and Neural Turing Machines  \citep{graves2014neural,graves2016hybrid}, exist, which all add to neural-network functionality, and which could all in-principle be trained in target space.  So in final conclusion, the target-space method seems to be a powerful additional tool which has tremendous potential for the enhancement of deep learning.

\newpage

\appendix

\section{Target-Space Algorithmic Complexity Calculations} \label{sec:computationComplexityCalculationsAppendix}
In this appendix we derive the algorithmic complexity for the main target-space algorithms. In these derivations, we ignore the computation of activation functions, and matrix additions, assuming these are dwarfed by matrix-multiplication operations.

\subsection{Computational Complexity for Fully-Connected Feed-forward networks} \label{sec:appendixComputationalComplexityFFNN}

First we consider the main target-space algorithm for feed-forward neural networks (i.e. Algorithms \ref{alg:mappingTargetsToWeightsSCU}-\ref{alg:layeredDiscreteFCCRealisationDEdT}).  

For a given layer $j$, the input matrix to that layer is $\layerInputStack{j}$, the weight matrix is $\layerWeightStack{j}$ and the target matrix is $T_j$.  For brevity, we will denote these three matrices without subscripts, as $A$, $W$ and $T$.  Let $\numinputs$ be an initialism for the  {\underline n}umber of {\underline i}nputs to the layer (i.e. the number of rows in $A$) and let $\numoutputs$ be the {\underline n}umber of {\underline o}utputs from the layer (i.e. the number of rows in $T$).  

Since $A \in \Reals^{\numinputs\times \nbt}$, and if $\nbt>\numinputs$, then direct multiplication to form the Gramian $AA^T$ will take ${\numinputs}^2\nbt$ floating-point operations (flops).  Assuming matrix inversion takes roughly $n^3$ flops, and since the Gramian is of shape $\numinputs\times \numinputs$, the formation of $(AA^T+\lambda I)^{-1}$ will take a further $(\numinputs)^3$ flops.  The formation of the product with $A^T$ in equation \eqref{eqn:BdaggerDefinition} will take a further $(\numinputs)^2\nbt$ flops.  Since $T \in \Reals^{\numoutputs \times \nbt}$, the multiplication by $T$ in equation \eqref{eqn:WlayeredDiscreteFCCRealisation} will take a further $\numinputs \numoutputs \nbt$ flops.  Hence summing these four terms gives the total time  to form the pseudoinverse and calculate the weight matrix in \eqref{eqn:WlayeredDiscreteFCCRealisation}, as $(\numinputs)^3+2(\numinputs)^2\nbt+\numinputs \numoutputs \nbt$.

If however $\nbt<\numinputs$, then the matrix $A$ is taller than it is wide, and \eqref{eqn:BdaggerDefinition} can be rearranged using the Woodbury matrix identity into an equivalent but more efficient form:
\begin{equation}
A^\pseudoinv \isDefinedToBe ( A^T{A}+\lambda I)^{-1}{A}^T. \label{eqn:BdaggerDefinitionAlternative}
\end{equation}
If this version is used, then the computational complexity is identically derived, resulting in the same flop-count expression but with all occurrences of $\numinputs$ and $\nbt$ swapped.  

Hence the resulting overall flop count for calculating $W$ by a pseudoinverse, assuming the faster of the two equations \eqref{eqn:BdaggerDefinition} and \eqref{eqn:BdaggerDefinitionAlternative} is used, is
\begin{equation}
    \text{Flop count for $W$ calculation}=\begin{cases}(\numinputs)^3+2(\numinputs)^2\nbt+\numinputs \numoutputs \nbt & \text{if $\numinputs < \nbt$}\\
    (\nbt)^3+2(\nbt)^2\numinputs+\numinputs \numoutputs \nbt& \text{otherwise}\end{cases}\label{eqn:flopcountWformation}
\end{equation}

Once the $W$ matrix for the layer is formed, the feed-forward calculation of the product $S_j=WA$ takes place, which is the same computational complexity as is required in ordinary weight space, i.e. requiring

\begin{equation}
    \text{Flop count for $S_j$ calculation}=\numinputs \numoutputs \nbt \label{eqn:flopcountSformation}
\end{equation}

If it can be assumed that the number of nodes in each layer of the neural network is approximately the same, so that $d_j=\overline{d}$ for all $j$, and no shortcut connections are present, then we can assume that $\numinputs\approx \numoutputs \approx \overline{d}$ (ignoring the single input from the bias node). 
If, as advocated in Section \ref{sec:minibatching}, we further assume that the size of the batch $\nbt$ is larger than $\overline{d}$ (so that also $\nbt>\numinputs$),  then summing the expressions in \eqref{eqn:flopcountWformation} and \eqref{eqn:flopcountSformation} and simplifying shows that the flop count for each layer of the target space Alg. \ref{alg:mappingTargetsToWeightsSCU} is bounded above by  $4\overline{d}^2\nbt$.  In comparison, the weight-space forward-pass algorithm for a single layer is just given by \eqref{eqn:flopcountSformation}, i.e. $\overline{d}^2\nb$ flops.  Hence the ratio of computation between target space and weight space is approximately upper-bounded by $(4\nbt/\nb)$.  Since automatic differentiation produces backward computations of similar algorithmic complexity as to the forward pass, the overall computation ratios for forward-and-backward passes between target space and weight space, when summed over all layers, is still approximately $(4\nbt/\nb)$.

\subsection{Computational Complexity for a CNN layer in Target Space} \label{sec:appendixCnnComputationalComplexity}

We now derive the computational complexity of the CNN target-space layer.  Notate the convolutional kernel width and heights by $\kernelwidth$ and $\kernelheight$ respectively, and the number of input and output channels by $\numinputchannels$ and $\numoutputchannels$ respectively.  Let $n_\mathrm{patch}$ be the number of image patches to be taken from each image.  Since the number of inputs operated on by the flattened $W$ matrix is $\numinputs=\kernelheight \kernelwidth \numinputchannels$, and the number of outputs is $\numoutputs=\numoutputchannels$, and the number of columns in the patches matrix $A$ is $n_\mathrm{b'}=\nbt n_\mathrm{patch}$, then substituting these factors into \eqref{eqn:flopcountWformation} gives a total flop count for the formation of $W$ as:
\begin{align}
\text{CNN Flop count for $W$ formation}&=
\begin{cases}(\kernelheight \kernelwidth \numinputchannels)^3+2(\kernelheight \kernelwidth \numinputchannels)^2n_\mathrm{b'}+\kernelheight \kernelwidth \numinputchannels \numoutputchannels n_\mathrm{b'} & \text{if $\kernelheight \kernelwidth \numinputchannels < n_\mathrm{b'}$}\\
    (n_\mathrm{b'})^3+2(n_\mathrm{b'})^2\kernelheight \kernelwidth \numinputchannels+\kernelheight \kernelwidth \numinputchannels \numoutputchannels n_\mathrm{b'}& \text{otherwise}
    \end{cases}\label{eqn:flopcountWformationCNN}
\end{align}

In contrast, the weight-space CNN forward pass only requires the formation of $S$, where the flop count is given by \eqref{eqn:flopcountSformation}, which equates to only $\kernelheight \kernelwidth \numinputchannels \numoutputchannels \nb n_\mathrm{patch}$ flops.

If we argue like in  Section \ref{sec:appendixComputationalComplexityFFNN} that $n_\mathrm{b'}>\numinputs$ (which is quite probable with the large number of image patches being processed by a CNN), and $\numoutputchannels\approx \numinputchannels$, then the flop count in target space is bounded above by 
\begin{align}
\text{CNN Flop count for $W$ formation}&\lessapprox3(\kernelheight \kernelwidth \numinputchannels)^2 n_\mathrm{b'}+\kernelheight \kernelwidth (\numinputchannels)^2 n_\mathrm{b'} \nonumber \\
&=(3(\kernelheight \kernelwidth)+1)\kernelheight \kernelwidth (\numinputchannels)^2 \nbt n_\mathrm{patch} \label{eqn:flopcountWformationCNNApproximation}
\end{align}
and hence the ratio of the flop count in target space to that in weight space is bounded above by approximately $(3(\kernelheight \kernelwidth)+1)\nbt/\nb$.

\section{Derivation of Algorithm \ref{alg:layeredDiscreteFCCRealisationDEdT}} \label{sec:alg:derivLayeredDiscreteFCCRealisationDEdT}

\subsection{Preliminary Definitions}

\noindent {\bf Single-Entry Matrix: }
\nobreak \\ 
Define $\Kij$ to be the {\it single-entry matrix}  with element at row $i$ and column $j$ equal to $\begin{cases}1&\text{if $m=i$ and $n=j$} \cr 0&\text{otherwise}\end{cases}$ \citep{petersen2012matrix}.  This is useful when
differentiating a matrix with respect to one of its elements, since $\fracpartial{A}{A^{ij}}=\Kij$, with  $\Kij$ having the same dimensions as $A$.

\noindent {\bf Raised Indices Notation: }
\nobreak \\ 
Define upper indices (without parentheses) after a matrix variable to indicate the matrix element, so that for example $A^{ij}$ is the element of $A$ with row index $i$ and column index $j$. 

Define raised indices $[ij]$ in square brackets after a scalar function $f(i,j)$ to mean the whole matrix whose element at row $i$ and column $j$ is $f(i,j)$.  For example, $\createMatrix{ij}{A^{ij}}\equiv A$, and $\createMatrix{ij}{A^{ji}}\equiv A^T$.

\noindent {\bf Frobenius Inner Product, $\frobProduct{A}{B}$}
\nobreak \\ 
For two $m \times n$ matrices $A$ and $B$, define $\frobProduct{A}{B}\isDefinedToBe \sum_{\forall i,j} A^{ij}B^{ij}$.  This inner product is useful when using the chain rule; for example, if $X$, $Y$ and $Z$ are matrices with $X=X(Y)$ and $Y=Y(Z)$ then
$\fracpartial{X^{mn}}{Z^{ij}}=\frobProduct{\fracpartial{X^{mn}}{Y}}{\fracpartial{Y}{Z^{ij}}}$.  Furthermore, if $\errorFunction(X)$ is a scalar function, then $\fracpartial{\errorFunction}{Y}=\createMatrix{ij}{\frobProduct{\fracpartial{X}{Y^{ij}}}{\fracpartial{\errorFunction}{X}}}$. 

\subsection{Basic Lemma for Combining Frobenius Inner Product with Single-entry Matrix}

A useful result for combining the inner product with $\Kij$ is 
\begin{equation}
\createMatrix{ij}{\frobProduct{A\Kij B}{C}}=A^TCB^T  \label{eqn:ruleHash}                      
\end{equation}
since $ \frobProduct{A\Kij B}{C}=\sum_{mn}(A\Kij B)^{mn}C^{mn} =\sum_{mn}\left(\sum_{pq}A^{mp}\Kij^{pq}B^{qn}\right)C^{mn}$

\noindent $=\sum_{mn}(A^{mi}B^{jn}C^{mn}) =(A^TCB^T)^{ij}$.

Similarly, \begin{equation}
          \createMatrix{ij}{\frobProduct{A\Kij^TB}{C}}=BC^TA 
          \label{eqn:ruleHashTranspose}
          \end{equation}

\subsection{Matrix Differentiation}
Differentiating a scalar by a matrix gives an identically dimensioned matrix,
e.g. $\left(\fracpartial{\errorFunction}{X}\right)^{ij}\isDefinedToBe \fracpartial{\errorFunction}{X^{ij}}$. 
Similarly for differentiating a matrix by a scalar:  
$\left(\fracpartial{X(a)}{a}\right)^{ij}\isDefinedToBe \fracpartial{X^{ij}(a)}{a}$.

Matrix differentiation follows the usual product rule:
\begin{align}
\fracpartial{AB}{X^{mn}}&=\fracpartial{A}{X^{mn}}B+A\fracpartial{B}{X^{mn}}.  \label{eqn:matrixProductRule1}
\end{align}

For example, if $A$,$B$ and $C$ are constant matrices, then
\begin{align*}
\fracpartial{AXBXC}{X^{mn}}&=A\Kmn BXC+AXB\Kmn C.
\end{align*} 

The derivative of an inverse matrix $A^{-1}$ is $\fracpartial{A^{-1}}{A^{ij}}=-A^{-1}\Kij A^{-1}$ \citep{brookes05}.  Combining this with the product rule gives 
\begin{align}
\fracpartial{(BB^T+\lambda I)^{-1}}{B^{ij}} 
& = -(BB^T+\lambda I)^{-1}\left(\Kij B^T+B\Kij^T\right)(BB^T+\lambda I)^{-1} \nonumber \\ 
& = -(BB^T+\lambda I)^{-1}\Kij B^\pseudoinv-(B^\pseudoinv)^T\Kij^T(BB^T+\lambda I)^{-1} 
\label{eqn:dBBinvDB}
\end{align}
And so,
\begin{align}
\fracpartial{B^\pseudoinv}{B^{ij}} =& \fracpartial{B^T(BB^T+\lambda I)^{-1}}{B^{ij}} &\text{(by \eqref{eqn:BdaggerDefinition})} \nonumber \\
=& \Kij^T(BB^T+\lambda I)^{-1} -B^T\fracpartial{(BB^T+\lambda I)^{-1}}{B^{ij}}&\text{(by product rule)}  \nonumber \\
=& \Kij^T(BB^T+\lambda I)^{-1} -\big(B^\pseudoinv \Kij B^\pseudoinv+B^\pseudoinv B\Kij^T(BB^T+\lambda I)^{-1}\big) & \hspace{-1cm}\text{(by \eqref{eqn:dBBinvDB})} \nonumber \\
=& (I-B^\pseudoinv B)\Kij^T(BB^T+\lambda I)^{-1}-B^\pseudoinv \Kij B^\pseudoinv   \label{eqn:dBDaggerDB}
\end{align}

\subsection{Ordered Partial Derivatives}\label{sec:orderedPartialDerivs}

Define the notation $\fracpartial{}{^*}$ to be the {\it ordered} partial derivatives \citep{werbos1974}, which take into account cascading changes to all later layers' weights and activations by Algorithm \ref{alg:mappingTargetsToWeightsSCU}.  For example  $\fracpartial{\vecW}{^*A_j^{mn}}$ describes how all the layers' weights would change according to Algorithm \ref{alg:mappingTargetsToWeightsSCU} if a small perturbation was forced to occur to $A_j^{mn}$.  

For a layer $j$, define
$\deltaStar{A_j} \isDefinedToBe \createMatrix{mn}{\fracpartial{\vecW}{^*A_j^{mn}}\dEdw}$.  
This matrix accounts for what effect a small change to
$A_j$ will have on $\errorFunction$, solely through the effect of cascading
changes to later layers' weights via alg.
\ref{alg:mappingTargetsToWeightsSCU}.  Note that $\deltaStar{A}$ is subtly  
different from $\fracpartial{\errorFunction}{^*A}$ since at the final layer
$\deltaStar{A_\nl}=0$ (since there are no later layers whose weights can
change), but
$\fracpartial{\errorFunction}{^*A_\nl}=\fracpartial{\errorFunction}{Y}\neq0$. Similarly, define 
$\deltaStar{S_j} \isDefinedToBe \createMatrix{mn}{\fracpartial{\vecW}{^*S_j^{mn}}\dEdw}$ and
$\deltaStar{\layerWeightStack{j}} \isDefinedToBe \createMatrix{mn}{\fracpartial{\vecW}{^*\layerWeightStack{j}^{mn}}\dEdw}$.

\subsection{Derivation of Algorithm \ref{alg:layeredDiscreteFCCRealisationDEdT}} 
Define $\deltaStar{\layerInputStack{j}}$ to be the composite of $\deltaStar{A_j}$ matrices in
the same way that the $\layerInputStack{j}$ matrices are composed of $A_j$ matrices, analogous to Eq. \eqref{eqn:BmatrixDefinition}.

The matrices $\layerWeightStack{j}$, $\layerInputStack{j}$, $T_j$, $Y_j$, $A_j$ and $S_j$ are for an arbitrary
layer $j$.  Throughout the following, all these matrices refer to the same subscripted value of $j$, therefore we omit this subscript to ease presentation.  To avoid the clash of variable names between $A_j$ and $\layerInputStack{j}$, we define $B \equiv \layerInputStack{j}$ and $A \equiv A_j$ as shorthand.

First we give useful results for $\fracpartial{W}{\Bmn}$ and $\deltaStar{W}$:
\begin{align}
\fracpartial{W}{\Bmn} =& T\big[(I-B^\pseudoinv B)\Kmn^T(BB^T+\lambda I)^{-1} -B^\pseudoinv \Kmn B^\pseudoinv \big]&\text{(by \eqref{eqn:WlayeredDiscreteFCCRealisation} and \eqref{eqn:dBDaggerDB})} \nonumber \\
=& (T-S)\Kmn ^T(BB^T+\lambda I)^{-1}-W\Kmn B^\pseudoinv  &\text{(by \eqref{eqn:WlayeredDiscreteFCCRealisation} and \eqref{eqn:S_ffnnEconomicalNotation})}  \label{eqn:proofLDFR:dWdB}
\end{align}

The derivation for $\deltaStar{W}=\createMatrix{mn}{\fracpartial{\vecW}{^*\layerWeightStack{j}^{mn}}\dEdw}$ in Equation \eqref{eqn:proofLDFR:deltaStarW} starts by adding two terms.  The first term, $\fracpartial{\errorFunction}{W}$, accounts for the contribution from the changing weights in that particular layer.  The second term,
$\createMatrix{pq}{\frobProduct{\deltaStar{S}}{\fracpartial{S}{W^{pq}}}}$, accounts for the cascading changes to all later layers' weights (by the definition of $\deltaStar{S}$).
\begin{align}
\deltaStar{W} =& 
	\fracpartial{\errorFunction}{W}+\createMatrix{mn}{\frobProduct{\deltaStar{S}}{\fracpartial{S}{W^{mn}}}}
	 && \nonumber \\
 =& \fracpartial{\errorFunction}{W}+\createMatrix{mn}{\frobProduct{\deltaStar{S}}{\Kmn B}}
	&\text{(by 
\eqref{eqn:S_ffnnEconomicalNotation} and \eqref{eqn:matrixProductRule1})}&\nonumber	\\
 =& 	\fracpartial{\errorFunction}{W}+(\deltaStar{S})B^T	 &\text{(by \eqref{eqn:ruleHash})}&
 \label{eqn:proofLDFR:deltaStarW}  
\end{align}

To derive a formula that calculates $\fracpartial{\errorFunctionT}{T}$ for a particular layer
given $\fracpartial{\errorFunction}{\vecW}$, we first note
that changing $T^{mn}$ for one layer will initially just change the weights of
that
layer, according to $\fracpartial{W}{T^{mn}}$.  Then cascading changes to the
later layers' weights will occur via Algorithm \ref{alg:mappingTargetsToWeightsSCU}, as a consequence of this initial
single layer's change of weights, and therefore all these cascading
effects are represented by $\deltaStar{W}$.  Combining these two factors with
the Frobenius inner-product gives:
\begin{align}
\fracpartial{\errorFunctionT}{T}=&  \createMatrix{mn}{\frobProduct{\deltaStar{W}}{\fracpartial{W}{T^{mn}}}} &&\nonumber\\
=&  \createMatrix{mn}{\frobProduct{\deltaStar{W}}{\Kmn B^\pseudoinv}} &\text{(by 
\eqref{eqn:WlayeredDiscreteFCCRealisation} and \eqref{eqn:matrixProductRule1})}& \nonumber\\
=&  \left( \fracpartial{\errorFunction}{W}+(\deltaStar{S}) B^T \right)(B^\pseudoinv)^T
&\text{(by \eqref{eqn:ruleHash} and \eqref{eqn:proofLDFR:deltaStarW})}& \label{eqn:proofLDFR:dEdT}
\end{align}

This requires calculation of the $\deltaStar{S}$ matrices for each layer.  
Since $A^{mn}=g(S^{mn})$, the chain rule gives 
\begin{equation}
 \deltaStar{S} = \hadamard{\deltaStar{A}}{g'(S)} \label{eqn:proofLDFR:deltaSToDeltaA}
\end{equation}

The derivation for $\deltaStar{B}$ is given in Equation \eqref{eqn:proofLDFR:dEdB}.  The first line of this derivation consists of two terms which are present, respectively, because changing $B$ will change the weights for that layer directly (via the equation $W=TB^\pseudoinv$), and will also change the sums for that layer directly (via the equation $S=WB$). The effects of these two changes are what the terms $\deltaStar{W}$ and $\deltaStar{S}$, respectively, are defined to represent.
\begin{align} 
\deltaStar{B}=& \createMatrix{mn}{\frobProduct{\deltaStar{W}}{\fracpartial{W}{B^{mn}}}
	+\frobProduct{\deltaStar{S}}{\fracpartial{S}{B^{mn}}}}&\nonumber\\
=& \createMatrix{mn}{\frobProduct{\deltaStar{W}}{\left((T-S)\Kmn ^T(BB^T+\lambda I)^{-1} - W\Kmn B^\pseudoinv \right)}+\frobProduct{\deltaStar{S}}{W \Kmn}} \hspace{-3cm}&\nonumber \\&&\text{(by \eqref{eqn:proofLDFR:dWdB}, \eqref{eqn:S_ffnnEconomicalNotation} and 
    \eqref{eqn:matrixProductRule1})}&\nonumber\\
=& (BB^T+\lambda I)^{-1}(\deltaStar{W})^T(T-S)-W^T(\deltaStar{W})(B^\pseudoinv)^T+W^T\deltaStar{S} 
	&\text{(by \eqref{eqn:ruleHash} and \eqref{eqn:ruleHashTranspose})}
	& \nonumber \\ 
=&W^T\left(\deltaStar{S}-\fracpartial{\errorFunctionT}{T}\right)
	+\bigg[(BB^T+\lambda I)^{-1}\left(\fracpartial{\errorFunction}{W}+(\deltaStar{S})B^T\right)^T(T-S)\bigg] &\text{(by
	\eqref{eqn:proofLDFR:dEdT} and \eqref{eqn:proofLDFR:deltaStarW})}\nonumber\\
	\label{eqn:proofLDFR:dEdB}
\end{align}
  
This enables us to find $\deltaStar{B}$ from $\deltaStar{S}$ for a particular
layer.  Since $\deltaStar{\layerInputStack{j}}\equiv \deltaStar{B}$ is
composed of $\deltaStar{A_{j-1}}$, and $\deltaStar{A_\nl}=0$ we can calculate the
$\deltaStar{A}$ matrices backwards, layer by layer.  Thus equations
\eqref{eqn:proofLDFR:dEdT}, \eqref{eqn:proofLDFR:deltaSToDeltaA} and
\eqref{eqn:proofLDFR:dEdB} give lines \ref{line:ldfrdedt:dEdTequals},
\ref{line:ldfrdedt:deltaSequals}, and \ref{line:ldfrdedt:deltaBequals} of Alg.
\ref{alg:layeredDiscreteFCCRealisationDEdT} respectively.

\section{Proof that a stationary point in target space corresponds to a stationary point in weight space}
\label{sec:appendix:stationaryPoints}
In this appendix we show that if $\optimal{\vecT}$ is a stationary point for the target-space problem, i.e. $\left.\dEdt\right|_{\vecT=\optimal{\vecT}}=0$, then the corresponding vector of weights $\optimal{\vecW}$ obtained from $\optimal{\vecT}$ through Algorithm \ref{alg:mappingTargetsToWeightsSCU} is a stationary point for the resulting weight-space problem, i.e. $\left.\dEdw\right|_{\vecW=\optimal{\vecW}}=0$.  

After a preliminary definition and two lemmas, the main theorem and proof follows.

\textbf{Definition:} Let
\begin{align}
    \qquad
    A^\pseudoinvinv \isDefinedToBe A + \lambda \, (A^\NRpseudoinv)^T.
    \label{eqn:pseudoinvinvDefinition}
\end{align}
where $A^{\NRpseudoinv}$ denotes the (non-regularised) Moore-Penrose pseudoinverse \cite[Section 5.5.2]{govl:13}.

\begin{lemma}For any real-valued matrix $A$, the following identity holds: $A\,A^{\pseudoinv} A^\pseudoinvinv = A$. \label{lem:reg_pseudoinv_identity}
\begin{proof}
We use the singular value decomposition (SVD) to prove this.  Let the shape of $A$ be $m\times n$, and the rank of $A$ be $r$.  Let the full SVD of $A$ be given by 
\begin{align}
    A=USV^T, \label{eqn:svdA}
\end{align} where the matrices $U\in \Reals^{m\times m}$, $S\in \Reals^{m\times n}$ and $V\in \Reals^{n\times n}$, the only non-zero elements of $S$ are on its leading diagonal, and where $U$ and $V$ are orthogonal.  Since $A$ is rank $r$, the matrix $S$ will have its first $r$ diagonal elements as non-zero and the remaining elements all zero.  Hence we can partition $S$ into block-matrix form as follows:
\begin{align}
    S=\begin{pmatrix}
    \Sigma & 0 \\ 0 & 0
    \end{pmatrix}, \label{eqn:SVD_s_blockmatrix_form}
    \end{align}
where $\Sigma$ is a diagonal matrix of shape $r\times r$, and the zeros are rectangular matrices of appropriate shape so as to make $S\in \Reals^{m \times n}$.  Since $\Sigma$ is square and full rank, its Moore-Penrose pseudoinverse simplifies into an ordinary inverse:
\begin{align}
   \Sigma^\NRpseudoinv=\Sigma^{-1}. \label{eqn:moorePenroseInverseForSigmaFullRank}
\end{align}

Using the SVD, and repeatedly cancelling orthogonal self-products such as $U^TU$ and $V^TV$, we can write:
\begin{align}
A^{\,\pseudoinv\,}&=VS^T(SS^T+\lambda I)^{-1}U^T &\text{(by \eqref{eqn:BdaggerDefinition})}\\    
A^{\NRpseudoinv}&=VS^{\NRpseudoinv}U^T\label{eqn:lem1Apsinv}&\text{(Moore-Penrose SVD)}\\
A^{\,\pseudoinvinv\,}&=U(S+\lambda S^{\NRpseudoinv})V^T&\text{(by \eqref{eqn:pseudoinvinvDefinition} and \eqref{eqn:lem1Apsinv})}
\end{align}
Substituting these into the left-hand side of the lemma's identity, and cancelling orthogonal self-products, gives,
\begin{align*}
 A\,A^{\pseudoinv} A^\pseudoinvinv &= USS^T(SS^T+\lambda I)^{-1}(S+\lambda S^{\NRpseudoinv})V^T   \\
 &= U\begin{pmatrix}
    \Sigma^2 & 0 \\ 0 & 0
    \end{pmatrix}\begin{pmatrix}
    (\Sigma^2+\lambda I)^{-1} & 0 \\ 0 & \lambda^{-1}I
    \end{pmatrix}\begin{pmatrix}
    \Sigma+\lambda \Sigma^{-1} & 0 \\ 0 & 0
    \end{pmatrix}V^T  &\text{(by \eqref{eqn:SVD_s_blockmatrix_form} and \eqref{eqn:moorePenroseInverseForSigmaFullRank})} \\
&= U\begin{pmatrix}
    \Sigma^2(\Sigma^2+\lambda I)^{-1}(\Sigma+\lambda \Sigma^{-1}) & 0 \\ 0 & 0
    \end{pmatrix}V^T   &\text{(block multiplication)}\\
&= U\begin{pmatrix}
    \Sigma(\Sigma^2+\lambda I)^{-1}(\Sigma^2+\lambda \Sigma \Sigma^{-1}) & 0 \\ 0 & 0
    \end{pmatrix}V^T &\hspace{-2cm}\text{(commute diagonal matrices)}\\
&= A   &\text{(by \eqref{eqn:svdA} and \eqref{eqn:SVD_s_blockmatrix_form})}
    \end{align*}
This proves the lemma.
\end{proof}
\end{lemma}
\textbf{Remark:} Lemma \ref{lem:reg_pseudoinv_identity} is analogous to the identity for the non-regularised Moore-Penrose pseudoinverse given by $A\,A^{\NRpseudoinv} A = A$  \cite[Section 5.5.2]{govl:13}, which also holds for any real-valued matrix $A$.

\begin{lemma}\label{lem:contradiction}
When the weights are calculated from the targets by Algorithm \ref{alg:mappingTargetsToWeightsSCU}, given any layer $j$, such that $j=\nl$ or $\fracpartial{\errorFunction}{\layerWeightStack{k}}=0$ for all $k>j$, we have $\fracpartial{\errorFunctionT}{T_j}=0 \Rightarrow \fracpartial{\errorFunction}{\layerWeightStack{j}}=0$.
\begin{proof}
First, let us write explicitly the derivative of the loss function $\errorFunction$ with respect to the weights of layer $j$:
\begin{align}
    \fracpartial{\errorFunction}{\layerWeightStack{j}} &= 
        \createMatrix{mn}{\frobProduct{\fracpartial{\errorFunction}{S_j}}
        {\fracpartial{S_j}{\layerWeightStack{j}^{mn}}}} \nonumber \\
        &= 
        \createMatrix{mn}{\frobProduct{\fracpartial{\errorFunction}{S_j}}
        {\Kmn \layerInputStack{j}}}
	&\text{(by 
\eqref{eqn:S_ffnnEconomicalNotation} and \eqref{eqn:matrixProductRule1})}
\nonumber \\
        &=\fracpartial{\errorFunction}{S_j}\,\layerInputStack{j}^T,&\text{(by \eqref{eqn:ruleHash})}
        \label{eqn:proof_convergenceW}
\end{align}

And similarly, explicitly state its derivative with respect to the targets of the same layer:
\begin{align}
    \fracpartial{\errorFunctionT}{T_j} &=\createMatrix{mn}{ \frobProduct{\fracpartial{\errorFunction}{S_j} +
            \sum_{k>j} \createMatrix{pq}{\frobProduct{\fracpartial{\errorFunction}{\layerWeightStack{k}}} {\fracpartial{\layerWeightStack{k}}{S_j^{pq}}}}
        }{
        \fracpartial{S_j}{T_j^{mn}}}}
        \hspace{-1.5em}.
        \label{eqn:proof_convergenceT}
\end{align}        
In this equation, instead of following \eqref{eqn:proofLDFR:dEdT}, we have used the chain rule to produce an expression that explicitly connects $\fracpartial{\errorFunctionT}{T}$ to $\fracpartial{\errorFunction}{S}$.  The summation in \eqref{eqn:proof_convergenceT} evaluates to $\deltaStar S_j$ defined in Section \ref{sec:orderedPartialDerivs}, which accounts for the effects of the weights in all later layers $k>j$ which will change by Alg. \ref{alg:mappingTargetsToWeightsSCU} as a result of a change to $T_j$.  Following on from the initial assumptions of this lemma, which were that either  $j=\nl$ or the condition $\fracpartial{\errorFunction}{\layerWeightStack{k}}=0$ holds for all $k>j$,
therefore the summation vanishes and \eqref{eqn:proof_convergenceT} reduces to:
\begin{align}
    \fracpartial{\errorFunctionT}{T_j} &=\createMatrix{mn}{ \frobProduct{\fracpartial{\errorFunction}{S_j}
        }{
        \fracpartial{S_j}{T_j^{mn}}}}& \nonumber \\
&=\createMatrix{mn}{ \frobProduct{\fracpartial{\errorFunction}{S_j}}{\fracpartial{\left(T_j{\left(\layerInputStack{j}\right)^\pseudoinv}\layerInputStack{j}\right)}{T_j^{mn}}}} &\text{(by \eqref{eqn:S_ffnnEconomicalNotation} and \eqref{eqn:WlayeredDiscreteFCCRealisation})}\nonumber \\        
&=\createMatrix{mn}{ \frobProduct{\fracpartial{\errorFunction}{S_j}}{\Kmn { \left(\layerInputStack{j}\right)^\pseudoinv}\layerInputStack{j} }} &\text{(by \eqref{eqn:matrixProductRule1})}\nonumber \\ 
 &=
\fracpartial{\errorFunction}{S_j} \, \layerInputStack{j}^T \,
\left(\layerInputStack{j}^T\right)^\pseudoinv. &\hspace{-8cm}\text{(by \eqref{eqn:ruleHash})}
\label{eqn:proof_convergenceTreduced}
\end{align}

Aiming from a contradiction, let us assume that $\fracpartial{\errorFunction}{\layerWeightStack{j}}\neq0$. Then one can choose an appropriately sized column vector $u$ such that $\fracpartial{\errorFunction}{\layerWeightStack{j}}\,u\neq0$. We now consider a second vector $v = \big(\layerInputStack{j}^T\big)^\pseudoinvinv u$, using \eqref{eqn:pseudoinvinvDefinition}, and write:
\begin{align}
    \fracpartial{\errorFunctionT}{T_j} \, v
    &=
        \fracpartial{\errorFunction}{S_j} \, \layerInputStack{j}^T \,
        \big(\layerInputStack{j}^T\big)^\pseudoinv \, v
        & \text{(by \eqref{eqn:proof_convergenceTreduced})} \label{eqn:proofContradiction1b}\\
    &=
        \fracpartial{\errorFunction}{S_j} \, \layerInputStack{j}^T \,
        \big(\layerInputStack{j}^T\big)^\pseudoinv \, \big(\layerInputStack{j}^T\big)^\pseudoinvinv \, u
        & \text{(by the definition of $v$)} \nonumber \\
    &=
        \fracpartial{\errorFunction}{S_j} \, \layerInputStack{j}^T \, u
        & \text{(by Lemma \ref{lem:reg_pseudoinv_identity})} \nonumber \\
    &=
        \fracpartial{\errorFunction}{\layerWeightStack{j}}\,u.
        & \text{(by \eqref{eqn:proof_convergenceW})} \label{eqn:proofContradiction4b}
\end{align}
While we initially assumed that the last line \eqref{eqn:proofContradiction4b} is non-zero, the first line \eqref{eqn:proofContradiction1b} must be zero, due to $\dEdt=0$. This contradiction proves the lemma.
\end{proof}
\end{lemma}

\begin{theorem}
When the weights are calculated from the targets by Algorithm \ref{alg:mappingTargetsToWeightsSCU}, we have $\dEdt=0 \implies \dEdw=0$.
\begin{proof}
We shall prove this result by induction, by first showing it holds for the last layer (used as the base case), and then showing that if it holds for all subsequent layers then it must also hold for the current layer (the inductive step).

Lemma \ref{lem:contradiction} explicitly handles the case where $j=\nl$, thus the base-case claim, that $\dEdt=0$ implies $\fracpartial{\errorFunction}{\layerWeightStack{\nl}}=0$, is true.  Next we consider the inductive step, i.e. that if $\dEdt=0$ and $\fracpartial{\errorFunction}{\layerWeightStack{k}}=0$ for all $k>j$, then $\fracpartial{\errorFunction}{\layerWeightStack{j}}$ must be zero.
Again, Lemma \ref{lem:contradiction} applies here, since it explicitly applies to $\fracpartial{\errorFunction}{\layerWeightStack{k}}=0$ for all $k>j$, and therefore the inductive step is also true.  This completes the proof by induction.
\end{proof}
\end{theorem}

This final theorem concludes the proof that $\dEdt=0$ implies $\dEdw=0$, i.e. that a stationary point for the target-space problem is also a stationary point for the corresponding weight-space problem obtained  through Algorithm \ref{alg:mappingTargetsToWeightsSCU}.

\end{document}